\documentclass[ijgi,article,accept,moreauthors,pdftex,10pt,a4paper]{mdpi} 

\usepackage{tummath}
\usepackage{subfigure}
\usepackage[noabbrev, capitalize]{cleveref}
\usepackage{csvsimple}
\usepackage{colortbl}
\usepackage{tabularx}
\usepackage[super]{nth}

\usepackage{tumabbrev} 

\usepackage{siunitx,booktabs}
\usepackage{contour}
\usepackage[acronym]{glossaries}

\usepackage{subfigure} 
\usepackage{tcolorbox}

\usepackage{xargs} 
\usepackage[colorinlistoftodos,prependcaption,textsize=small]{todonotes}

\usepackage{pgfplots}
\usepgfplotslibrary{dateplot}

\firstpage{1}
\pubvolume{7}
\issuenum{4}
\articlenumber{129}
\pubyear{2018}
\copyrightyear{2018}
\history{Received: 22 January 2018; Accepted: 17 March 2018; Published: 21 March 2018; Updated on ArXiv: 7 April 2018}

\pdfoutput=1

\tikzset{
	mylabel/.style={
		label={
			[label distance=-1.5em]north:
			\contour{white}{
				\small #1
			}
		}
	}
}

\tikzset{
	mylabel2/.style={
		label={
			[label distance=-1.75em]north:
			\contour{white}{
				\small #1
			}
		}
	}
}

\usepackage{booktabs}
\usepackage{array}
\newcolumntype{x}{l}
\newcolumntype{X}{>{\scriptsize}l}
\newcolumntype{v}[1]{>{\raggedright\hspace{0pt}}p{#1}}
\newcolumntype{V}[1]{>{\scriptsize\raggedright\hspace{0pt}}p{#1}}

\newcommand{\best}[1]{\textbf{\color{tumblack}#1}}
\newcommand{\worst}[1]{\textbf{\color{tumblack}#1}}
\newcommand{\tabmetricsbothtex}[1]{
  
  \begin{tabular}{@{}Xxxxxxxxxxx@{}}%
    \toprule
    \multicolumn{1}{@{}X}{Class} & \multicolumn{8}{X@{}}{Year} \\
    \cmidrule(l){2-11} 
    & \multicolumn{5}{X}{2016} & \multicolumn{5}{X@{}}{2017} \\
    \cmidrule(lr){2-6}
    \cmidrule(l){7-11} 
    &
    \multicolumn{1}{X}{Precision} &
    \multicolumn{1}{X}{Recall} &
    \multicolumn{1}{X}{$f$-Meas.} &
    \multicolumn{1}{X}{Kappa} &
    \multicolumn{1}{X}{\#Pixels} &
    \multicolumn{1}{X}{Precision} &
    \multicolumn{1}{X}{Recall} &
    \multicolumn{1}{X}{$f$-Meas.} &
    \multicolumn{1}{X}{Kappa} &
    \multicolumn{1}{X}{\#Pixels} \\
    
    &
    \multicolumn{1}{X}{(Users Acc.)} &
    \multicolumn{1}{X}{(Prod. Acc.)} &
    \multicolumn{1}{X}{} &
    \multicolumn{1}{X}{} &
    \multicolumn{1}{X}{} &
    \multicolumn{1}{X}{(Users Acc.)} &
    \multicolumn{1}{X}{(Prod. Acc.)} &
    \multicolumn{1}{X}{} &
    \multicolumn{1}{X}{} & \\
    \cmidrule(r){1-1}
    \cmidrule(lr){2-2}
    \cmidrule(lr){3-3}
    \cmidrule(lr){4-4}
    \cmidrule(lr){5-5}
    \cmidrule(lr){6-6}
    \cmidrule(lr){7-7}
    \cmidrule(lr){8-8}
    \cmidrule(lr){9-9}
    \cmidrule(lr){10-10}
    \cmidrule(l){11-11}
    \addlinespace
    
    \cn{sugar beet}	     &	94.6    &	77.6		&	85.3		& .772 			&	59k		&	89.2		&	78.5		&	83.5		& .779 &	94k	\\
    \cn{oat}	     &	86.1    &	67.8		&	75.8		& .675 			&	36k		&	63.8		&	62.8		&	63.3		& .623 &	38k	\\
    \cn{meadow}		     &	90.8    &	85.7		&	88.2		& .845 			&	233k	&	88.1		&	85.0		&	86.5		& .837 &	242k	\\
    \cn{rapeseed}		     &	95.4    &	90.0		&	92.6		& .896 			&	125k	&\best{96.2}	&	95.9		&\best{96.1}	& \best{.957} &	114k	\\
    \cn{hop}		     &\best{96.4}&	87.5		&	91.7		& .873 			&	51k		&	92.5		&	74.7		&	82.7		& .743 &	53k	\\
    \cn{spelt}	 &\worst{55.1}&	81.1		&	65.6		& .807 			&	38k		&	75.3		&	46.7		&	57.6		& .463 &	31k	\\
    \cn{triticale}&	69.4	&	55.7		&	61.8		& .549 			&	65k		&	62.4		&	57.2		&	59.7		& .563 &	64k	\\
    \cn{beans}		     &	92.4	&	87.1		&	89.6		& .869 			&	27k		&	92.8		&	63.2		&	75.2		& .630 &	28k	\\
    \cn{peas}		     &	93.2	&	70.7		&	80.4		& .706 			&	9k		&\worst{60.9}	&\worst{41.5}	&\worst{49.3}	& \worst{.414} & 	6k	\\
    \cn{potato}		     &	90.9	&	88.2		&	89.5		& .876 			&	126k	&	95.2		&	73.8		&	83.1		& .728 &	140k	\\
    \cn{soybeans}	     &	97.7	&	79.6		&	87.7		& .795 			&	21k		&	75.9		&	79.9		&	77.8		& .798 &	26k	\\
    \cn{asparagus}	     &	89.2	&	78.8		&	83.7		& .787 			&	20k		&	81.6		&	77.5		&	79.5		& .773 &	19k	\\
    \cn{wheat}	 &	87.7	&	93.1		&	90.3		& .902 			&	806k	&	90.1		&	95.0		&	92.5		& .930 &	783k	\\
    \cn{winter barley}	 &	95.2	&	87.3		&	91.0		& .861 			&	258k	&	92.5		&	92.2		&	92.4		& .915 &	255k	\\
    \cn{rye}	     &	85.6	&\worst{47.0}	&\worst{60.7}	& \worst{.466} 	&	43k		&	76.7		&	61.9		&	68.5		& .616 &	30k	\\
    \cn{summer barley}	 &	87.5	&	83.4		&	85.4		& .830 			&	73k		&	77.9		&	88.5		&	82.9		& .880 &	91k	\\
    \cn{maize}		     &	91.6	&\best{96.3}	&\best{93.9}	& \best{.944} 	&	919k	&	92.3		&\best{96.8}	&	94.5		& .953 &	876k	\\
    \addlinespace
    \textbf{weight. avg}	&\textbf{89.9}	&\textbf{89.7}	&\textbf{89.5}	& \textbf{} &	&\textbf{89.5}	&\textbf{89.5}	&\textbf{89.3}	& \textbf{} &		\\
    \cmidrule(lr){2-6}
    \cmidrule(l){7-11} 
    
    \addlinespace
    & \multicolumn{2}{X}{Overall Accuracy} & \multicolumn{2}{X}{Overall Kappa} & & \multicolumn{2}{X}{Overall Accuracy} & \multicolumn{2}{X}{Overall Kappa} \\
    \cmidrule(lr){2-3}
    \cmidrule(lr){4-5}
    \cmidrule(lr){7-8}
    \cmidrule(lr){9-10}
    & \textbf{89.7} & & \textbf{.870} & & & \textbf{89.5} & & \textbf{.870} & \\

    \addlinespace
    \bottomrule
  \end{tabular}
}

\newcommand{\tabformulas}{

\begin{table}
  \centering
  \glsreset{rnn}
  \glsreset{lstm}
  \glsreset{gru}
  \caption{Update formulas of the convolutional variants of standard \glspl{rnn}, \gls{lstm} cells and \glspl{gru}.
    A convolution between matrices $\V{a}$ and $\V{b}$ is denoted by $\conv{\V{a}}{\V{b}}$, element-wise multiplication by the \emph{Hadamard operator} $\V{a} \odot \V{b}$, and concatenation on the last dimension is marked by $\concat{\V{a}}{\V{b}}$.
    The activation functions sigmoid $\sigma(x)$ and tangens hyperbolicus $\operatorname{tanh}(x)$ are used for non-linear scaling. 
  }
  \label{tab:rnn}
  
    \begin{tabular}{@{}xxxx@{}}
      \toprule
      \multicolumn{1}{@{}X}{Gate} & \multicolumn{3}{X@{}}{Variant} \\
      \cmidrule(l){2-4}   
       & \multicolumn{1}{X}{\gls{rnn}} 
       & \multicolumn{1}{X}{\acrshort{lstm} \cite{Hochreiter97:LST}} 
       & \multicolumn{1}{X@{}}{\acrshort{gru} \cite{Cho2014}} \\
      \cmidrule(r){1-1} 
      \cmidrule(lr){2-2}
      \cmidrule(lr){3-3}
      \cmidrule(l){4-4}  
      \addlinespace
      \multicolumn{1}{@{}X}{} & 
      $\VHidden_t \leftarrow \VInput_t, \VHidden_{t-1}$ & 
      $\VHidden_t,\VCellState_t  \leftarrow \VInput_t, \VHidden_{t-1}, \VCellState_{t-1}$& 
      $\VHidden_t \leftarrow\VInput_t, \VHidden_{t-1}$ \\
      \addlinespace
      \multicolumn{1}{@{}X}{Forget/Reset} & 
      & 
      $\VForgetGate_t \leftarrow \sigma( \conv{\concat{ \VInput_t }{ \VHidden_{t-1} } }{ \MWeight_f } + 1 ) $ & 
      $\VResetGate_t \leftarrow \sigma( \conv{ \concat{ \VInput_t }{ \VHidden_{t-1} } }{ \MWeight_r } )$ \\
      \addlinespace
      \multicolumn{1}{@{}X}{Insert/Update} & 
      & 
      $\VInputGate_t \leftarrow \sigma( \conv{ \concat{ \VInput_t }{ \VHidden_{t-1} } }{ \MWeight_i } ) $ &
      $\VUpdateGate_t \leftarrow \sigma( \conv{ \concat{ \VInput_t }{ \VHidden_{t-1} } }{ \MWeight_u } )$ \\
      
      & & $\VModulationGate_t \leftarrow \sigma( \conv{ \concat{ \VInput_t }{ \VHidden_{t-1} } }{ \MWeight_j } ) $ & \\
      \addlinespace
      \multicolumn{1}{@{}X}{Output} &  
      & 
      $\VOutputGate_t \leftarrow \sigma( \conv{ \concat{ \VInput_t }{ \VHidden_{t-1} } }{ \MWeight_o} ) $ & $\tilde{\VHidden}_t \leftarrow  \conv{ \concat{ \VInput_t }{ \VResetGate_t \odot \VHidden_{t-1} } }{ \MWeight_{\tilde{h}} }$ \\
      \addlinespace
      \multicolumn{1}{@{}X}{} & &
      $\VCellState_t \leftarrow \VCellState_{t-1} \odot \VForgetGate_t + \VInputGate_t \odot \VModulationGate_t$ & \\
      \addlinespace
      
      \multicolumn{1}{@{}X}{} & 
      $\VHidden_t \leftarrow \sigma( \conv{ \concat{ \VInput_t }{ \VHidden_{t-1} } }{ \MWeight })$ & 
      $\VHidden_t \leftarrow \VOutputGate_t \odot \tanh(\VCellState_t) $ &
      $ \VHidden_t \leftarrow \VUpdateGate_t \odot \VHidden_{t-1} + (1 - \VUpdateGate_t) \odot \tanh( \tilde{\VHidden}_t ) $ \\
      \addlinespace
      \bottomrule
    \end{tabular}
\end{table}
  }
  
\newcommand{\tabAccuracies}{
\begin{table}
  \centering
  \caption{Pixel-wise accuracies of the trained convolutional \gls{gru} sequential encoder network after training over 60 epochs on data of both growth seasons. The conditional kappa metrics \citep{Fung1988} for each class and the overall kappa \citep{cohen1960} measure are given for both growth seasons.
  }
  \label{tab:accuracies}
  
  \tabmetricsbothtex{tables/gru256.tex}
\end{table}
}

\newacronym{eo}{EO}{Earth observation} 
\newacronym{lulc}{LULC}{land use and land cover classification}
\newacronym{gpu}{GPU}{graphics processing unit}
\newacronym{ndvi}{NDVI}{normalized difference vegetation index}
\newacronym{ndwi}{NDWI}{normalized difference water index}
\newacronym{evi}{EVI}{enhanced vegetation index}
\newacronym{vi}{VI}{vegetation index}
\newacronym{obia}{OBIA}{object based image analysis}
\newacronym{dt}{DT}{decision tree}
\newacronym{rf}{RF}{random forest}
\newacronym{hmm}{HMM}{hidden Markov model}
\newacronym{crf}{CRF}{conditional random field}
\newacronym{rnn}{RNN}{recurrent neural network}
\newacronym{lstm}{LSTM}{long short-term memory}
\newacronym{dl}{DL}{deep learning}
\newacronym{gru}{GRU}{gated recurrent unit}
\newacronym{aoi}{AOI}{area of interest}
\newacronym{aoipl}{AOIs}{areas of interest}
\newacronym{svm}{SVM}{support vector machine}
\newacronym{stmelf}{StMELF}{Bavarian Ministry of Food, Agriculture and Forestry}
\newacronym{toa}{TOA}{top-of-atmosphere}
\newacronym{boa}{BOA}{bottom-of-atmosphere}
\newacronym{convrnn}{ConvRNN}{convolutional recurrent neural network}
\newacronym{cnn}{CNN}{convolutional neural network}
\newacronym{gsd}{GSD}{ground sampling distance}
\newacronym{relu}{ReLU}{rectified linear unit}
\newacronym{ann}{ANN}{artificial neural network}
\newacronym{ffn}{FNN}{feed-forward network}
\newacronym{nlp}{NLP}{natural language processing}
\newacronym{crs}{CRS}{coordinate reference system}
\newacronym{lrz}{LRZ}{Leibnitz Supercomputing Centre}
\newacronym{idlstm}{ID-LSTM}{Incremental Dual-memory LSTM}
\newacronym{s2}{S2}{Sentinel 2}
\newacronym{ls}{LS}{Landsat}
\newacronym{modis}{MODIS}{Moderate-resolution Imaging Spectroradiometer}
\newacronym{aster}{ASTER}{Advanced Spaceborne Thermal Emission and Reflection Radiometer}
\newacronym{spot}{SPOT}{Satellite Pour l’Observation de la Terre}
\newacronym{rapideye}{RE}{RapidEye}
\newacronym{quickbird}{QB}{QuickBird}

\makeatletter
\newcommand*{\glsplainhyperlink}[2]{%
  \colorlet{currenttext}{.}
  \colorlet{currentlink}{\@linkcolor}
  \hypersetup{linkcolor=currenttext}
  \hyperlink{#1}{#2}%
  \hypersetup{linkcolor=currentlink}
}
\let\@glslink\glsplainhyperlink
\makeatother

\defglsdisplayfirst[acronym]{{#1#4}} 
\defglsdisplayfirst{{#1#4}}

\usetikzlibrary{3d}

\usepackage{pgfplotstable}
\usepgfplotslibrary{external}
\tikzsetexternalprefix{tikz/}

\usepackage{pdfpages}

\tikzstyle{perspective}=[every node/.append style={
  yslant=-.2,xslant=1},yslant=0.2,xslant=1]

\tikzstyle{perspective3d}=[
x={(0.5cm,0.5cm)}, y={(1cm,0cm)}, z={(0cm,1cm)}]

\usepackage{tumtensors} 

\usepackage{tumcolors}

\usetikzlibrary{matrix}

\DeclareSIUnit[
product-units=single
]\pixel{px}
\DeclareSIUnit{\nothing}{\relax}
\usetikzlibrary{arrows} 
\usetikzlibrary{backgrounds}
\usetikzlibrary{fit}
\usetikzlibrary{shapes}
\usetikzlibrary{calc}
\usetikzlibrary{positioning}
\usepackage{mwe}
\usetikzlibrary{3d,calc}
\usepackage[nodayofweek,level]{datetime}

\usepackage{lscape}

\usepackage{pgfplots}
\usepgfplotslibrary{groupplots}

\colorlet{traincolor}{tumbluelight}
\colorlet{validcolor}{tumbluedark}
\colorlet{evalcolor}{tumorange}

\newcommand{\newtext}[1]{{#1}}

\newcommand{\classname}[1]{\textsl{#1}} 
\newcommand{\cn}[1]{\classname{#1}} 

\newcommand{\brand}[1]{\textsc{#1}}

\newcommand{\satellite}[1]{\brand{#1}}
\newcommand{\band}[1]{\brand{#1}}

\newcommand{\MWeight}{\ensuremath{\M{W}}}

\newcommand{\VInput}{\DataVec}
\newcommand{\VHidden}{\ensuremath{\V{h}}}

\newcommand{\VCellState}{\ensuremath{\V{c}}}
\newcommand{\VForgetGate}{\ensuremath{\V{f}}}
\newcommand{\VModulationGate}{\ensuremath{\V{j}}}
\newcommand{\VInputGate}{\ensuremath{\V{i}}}
\newcommand{\VOutputGate}{\ensuremath{\V{o}}}

\newcommand{\VResetGate}{\ensuremath{\V{r}}}
\newcommand{\VUpdateGate}{\ensuremath{\V{u}}}

\newcommand{\concat}[2]{[#1 \Vert #2]}

\newcommand{\Rin}[3]{\mathbb{R}^{#1 \times #2 \times #3}}
\newcommand{\Rinfloor}[3]{\mathbb{R}^{#1} \times \mathbb{R}^{#2} \times \mathbb{R}^{#3}}
\newcommand{\Rinfour}[4]{\mathbb{R}^{#1 \times #2 \times #3 \times #4}}

\newcommand{\krnn}{k_\text{rnn}}
\newcommand{\kclass}{k_\text{class}}

\usepackage{array}
\newcolumntype{L}[1]{>{\raggedright\let\newline\\\arraybackslash\hspace{0pt}}m{#1}}
\newcolumntype{C}[1]{>{\centering\let\newline\\\arraybackslash\hspace{0pt}}m{#1}}
\newcolumntype{R}[1]{>{\raggedleft\let\newline\\\arraybackslash\hspace{0pt}}m{#1}}

\usepackage{setspace}

\Title{Multi-Temporal Land Cover Classification with Sequential Recurrent Encoders}

\Author{Marc Ru{\ss}wurm *\orcidA{} and Marco Körner \orcidB{}}

\AuthorNames{Marc Ru\ss{}wurm and Marco K\"orner}

\address{%
Chair of Remote Sensing Technology, TUM Department of Civil, Geo and Environmental Engineering, Technical University of Munich, Arcisstraße 21, 80333 Munich, Germany; marco.koerner@tum.de\\
}

\corres{\hangafter=1 \hangindent=1.05em \hspace{-0.82em} Correspondence: marc.russwurm@tum.de; Tel.: +49-172-81-70-121}

\abstract{
   \Gls{eo} sensors deliver data at daily or weekly intervals.
   Most \gls{lulc} approaches, however, are designed for cloud-free and mono-temporal observations.
   The increasing temporal capabilities of today's sensors enable the use of temporal, along with spectral and spatial features.
   Domains such as speech recognition or neural machine translation, work with inherently temporal data and, today, achieve impressive results by using sequential encoder-decoder structures.
   Inspired by these sequence-to-sequence models, we adapt an encoder structure with convolutional recurrent layers in order to approximate a phenological model for vegetation classes based on a temporal sequence of \gls{s2} images. 
   In our experiments, we visualize internal activations over a sequence of cloudy and non-cloudy images and find several recurrent cells that reduce the input activity for cloudy observations.
   Hence, we assume that our network has learned cloud-filtering schemes solely from input data, which could alleviate the need for tedious cloud-filtering as a preprocessing step for many \gls{eo} approaches.
   Moreover, using unfiltered temporal series of \gls{toa} reflectance data, our experiments achieved state-of-the-art classification accuracies on a large number of crop classes with minimal preprocessing, compared to other classification approaches.
  }


\begin{document}
%
%
%

\section{Introduction}
\label{sec:introduction}

\glsresetall 

\Gls{lulc} has been a central focus of \gls{eo} since the first air- and space-borne sensors began to provide data.
For this purpose, optical sensors sample the spectral reflectivity of objects on the Earth's surface in a spatial grid at repeated intervals.
Hence, \gls{lulc}~classes can be characterized by spectral, spatial and temporal features.
Today, most~classification tasks focus on spatial and spectral features~\citep{Zhang18}, while utilizing the temporal domain had long proven challenging.
This is mostly due to limitations on data availability, the cost of data acquisition, infrastructural challenges regarding data storage and processing and the complexity of model design and feature extraction over multiple time frames.

Some \gls{lulc} classes, such as urban structures, are mostly invariant to temporal changes and, hence, are suitable for mono-temporal approaches.
Others, predominantly vegetation-related classes, change~their spectral reflectivity based on biochemical processes initiated by {phenological} events related to the type of vegetation and to environmental conditions.
These~vegetation-characteristic phenological transitions have been utilized for crop yield prediction and, to some extent, for~classification~\citep{odenweller1984,Reed1994}.
However, to circumvent the previously-mentioned challenges, the~dimensionality of spectral bands has often been compressed by calculating task-specific indices, such as the {normalized difference vegetation index (NDVI)}, the {normalized difference water index (NDWI)} or the {enhanced vegetation index (EVI)}.

Today, most of these temporal data limitations have been alleviated by technological advances.
Reasonable spatial and temporal resolution data of multi-spectral Earth observation sensors are available at no cost.
Moreover, new services inexpensively provide high temporal and spatial resolution imagery.
The cost of data storage has decreased, and data transmission has become sufficiently fast to allow gathering and processing all available images over a large area and multiple years.
Finally, new~advances in machine learning, accompanied by GPU-accelerated hardware, have made it possible to learn complex functional relationships, solely from the data provided.

Since now data are available at high resolutions and processing is feasible, the temporal domain should be exploited for \gls{eo} approaches.
However, this exploitation requires suitable processing techniques utilizing all available temporal information at reasonable complexity.
Other domains, such~as machine translation~\citep{Bahdanau14}, text summarization~\citep{Rush15, Shen16, Nallapati16} or speech recognition~\citep{Sutskever14, Chorowski15}, handle sequential data naturally.
These domains have popularized {sequence-to-sequence learning}, which transforms a variable-length input sequence to an intermediate representation.
This representation is then decoded to a variable-length output sequence.
From this concept, we adopt the sequential encoder structure and extract characteristic temporal features from a sequence of \gls{s2} images using a straightforward, two-layer network.


Thus, the main contributions of this work are:
\begin{enumerate}[align=parleft,leftmargin=10mm, labelsep=3mm]
	\item[(i)] the adaptation of sequence encoders from the field of sequence-to-sequence learning to \acrfull{eo},
	\item[(ii)] a visualization of internal gate activations on a sequence of satellite observations and,
	\item[(iii)] the application of crop classification over two seasons.
\end{enumerate}

\section{Related Work}
\label{sec:relatedwork}

As we aim to apply our network to vegetation classes, we first introduce common crop classification approaches, to which we will compare our results in \cref{sec:discussion}.
Then, we motivate data-driven learning models and cover the latest work on recurrent network structures in the \gls{eo}~domain.

\textls[-20]{Many remote sensing approaches have achieved adequate classification accuracies for multi-temporal crop data by using multiple preprocessing steps in order to improve feature separability.
	Common~methods are atmospheric correction~\cite{Foerster2012, conrad2010, conrad2014, Hao15, Barragan2011}, calculation of vegetation indices~\cite{Foerster2012, conrad2010, conrad2014, Hao15, Barragan2011} or~the extraction of sophisticated phenological features~\cite{Hao15}.
	Additionally, some approaches utilize expert knowledge, for~instance, by introducing additional agro-meteorological data~\cite{Foerster2012}, by selecting suitable observation dates for the target crop-classes~\cite{Barragan2011} or by determining rules for classification~\cite{conrad2010}.
	Pixel-based~\citep{Foerster2012, Hao15} and object-based~\citep{conrad2010,conrad2014,Barragan2011} approaches have been proposed.}
Commonly, \glspl{dt}~\citep{Foerster2012, conrad2010, Barragan2011} or \glspl{rf}~\citep{conrad2014, Hao15} are used as classifiers, the rules of which are sometimes aided by additional expert knowledge~\cite{conrad2010}.

These traditional approaches generally trade procedural complexity and the use of region-specific expert knowledge for good classification accuracies in the respective \gls{aoipl}.
\glsunset{aoi} 
However, these approaches are, in general, difficult to apply to other regions.
Furthermore, the processing structure requires supervision to varying degrees (e.g., product selection, visual image inspection, parameter tuning), which impedes application at larger scales.

Today, we are experiencing a change in paradigm: away from the design of physically-interpretable, human-understandable models, which require task-specific expert knowledge, towards~data-driven models, which are encoded in internal weight parameters and derived solely from observations.
In that regard, \glspl{hmm}~\cite{Siachalou2015} and \glspl{crf}~\cite{Hoberg2015:CRF} have shown promising classification accuracies with multi-temporal data.
However, the underlying {Markov property} limits long-term learning capabilities, as Markov-based approaches assume that the present state only depends on the current input and {one} previous state.

{
	{Deep learning} methods have had major success in fields, such as target recognition and scene understanding~\cite{Zhang16}, and are increasingly adopted by the remote sensing community.
	These} methods have proven particularly beneficial for modeling physical relationships that are complicated, cannot~be generalized or are not well-understood~\citep{Zhu2017}.
Thus, deep learning is potentially well suited to approximate models of phenological changes, which depend on complex internal biochemical processes of which only the change of surface reflectivity can be observed by \gls{eo} sensors.
A purely data-driven approach might alleviate the need to manually design a functional model for this complex relationship.
However, caution is required, as external and non class-relevant factors, such as seasonal weather or observation configurations, are potentially incorporated into the model, which might remain undetected if these factors constantly bias the dataset.

In remote sensing, {convolutional networks} have gained increasing popularity for mono-temporal observation tasks~\citep{Hu15:TDC,Scott17:TDC,Makantasis2015,Castelluccio2015}.
However, for sequential tasks, recurrent network architectures, which~provide an iterative framework to process sequential information, are generally better suited.
{Recent approaches utilize recurrent architectures for change detection~\citep{Lyu2016, Jia2017b, Lichao2018}}, identification of sea level anomalies~\citep{Braakmann-Folgmann2017} and land cover classification~\citep{Sharma17}.
For long-term dependencies,~\citet{Jia2017b} proposed a new cell architecture, which maintains two separate cell states for single- and multi-seasonal long-term dependencies.
However, the calculation of an additional cell state requires more weights, which may prolong training and require more training samples.

In previous work, we have experimented with recurrent networks for crop classification~\cite{Russwurm17:TVM} and achieved promising results.
Based on this, we propose a network structure using convolutional recurrent layers and the aforementioned adaptation of a {many-to-one} classification scheme with sequence encoders.

\section{Methodology}
\label{sec:methodology}

\cref{sec:seqenc} incrementally introduces the concepts of \glspl{ann}, \glspl{ffn}, and \glspl{rnn} and illustrates the use of \glspl{rnn} in sequence-to-sequence learning.
We then describe details of the proposed network structure in \cref{sec:approach}.
  
\subsection{Network Architectures and Sequential Encoders}
\label{sec:seqenc}

\tabformulas

\begin{figure}
  \centering
  \tikzsetnextfilename{lstm}

\tikzstyle{operator} = [draw, circle, fill=tumbluemedium, draw=tumbluemedium, inner sep=0, text=white]
\tikzstyle{gate} = [fill=tumivory,draw,rounded corners, minimum height=1.7em]

\tikzstyle{dummy} = [inner sep=0]
\tikzstyle{flow} = [rounded corners]
\tikzstyle{endflow} = [-stealth,flow]

\colorlet{boxcolor}{tumgraylight}
\tikzstyle{bigbox} = [rectangle, draw=tumivory, thick, fill=boxcolor, rounded corners, 
inner sep=2ex]

\tikzset{pic shift/.store in=\shiftcoord,
	pic shift={(0,0)},
	lstm/.pic = {
		\begin{scope}[shift={\shiftcoord},xscale=3,yscale=2]
			
			\node[dummy] (bl) at (0,0){}; 
			\node[dummy] (tr) at (1,1){}; 
			
			\node[dummy] (br) at ($ (bl -| tr) $){}; 
			\node[dummy] (tl) at ($ (bl |- tr) $){}; 
			
			\node[fit=(bl) (tr),bigbox] (-C) {};
			
			\coordinate (-input) at (0.1,1); 
			
			\coordinate (-coutput) at (0.9,0); 
			\coordinate (-houtput) at (0.9,1); 
			
			\def\d{1/6}
			
			\def\h{1/3}
			
			\coordinate (f)  at bl+(1*\d,0);
			\coordinate (i)  at bl+(2*\d,0);
			\coordinate (j)  at bl+(3*\d,0);
			\coordinate (o)  at bl+(4*\d,0);
			\coordinate (out) at bl+(5*\d,0);
			
			\coordinate (gates) at (0,2*\h);
			
			
			
			\node[gate](fgate) at ($ (gates -| f) $){$\VForgetGate_t$};
			\node[gate](igate) at ($ (gates -| i) $){$\VInputGate_t$};
			\node[gate](jgate) at ($ (gates -| j) $){$\VModulationGate_t$};
			\node[gate](ogate) at ($ (gates -| o) $){$\VOutputGate_t$};
			
%
			\node[operator](fmult) at ($ (bl -| fgate) $) {$ \odot $};
			\draw[endflow] (-input) -| (fgate) -- (fmult); 
			
			\node[operator](jmult) at ([shift={(0,-1*\h)}]jgate) {$ \odot $};
			\node[operator](cadd) at ($ (bl -| jgate) $) {$ + $};
			\draw[endflow] (-input) -| (jgate) -- (jmult);
			\draw[endflow] (jmult) -- (cadd); 			

			\draw[endflow] (-input) -| (igate) |- (jmult); 
%
			\node[operator](outtanh) at ([shift={(0,1*\h)}]out) {$\odot$};
%
			\draw[endflow] (tl) -| (ogate) |- (outtanh);
			\draw[flow] (outtanh) |- (-houtput);
%
			\draw[endflow] (cadd) -| (outtanh);
			\draw[flow] (fmult) -- (cadd) -- (-coutput);
%

			
		\end{scope}
	}
}

\tikzset{pic shift/.store in=\shiftcoord,
	pic shift={(0,0)},
	gru/.pic = {
		\begin{scope}[shift={\shiftcoord},xscale=3,yscale=2]
			
			\node[dummy] (bl) at (0,0){}; 
			\node[dummy] (tr) at (1,1){}; 
			
			\node[dummy] (br) at ($ (bl -| tr) $){}; 
			\node[dummy] (tl) at ($ (bl |- tr) $){}; 
			
			\node[fit=(bl) (tr),bigbox] (-C) {};
			
			\def\d{1/5}
			
			\def\h{1/4}

			\coordinate (-xinput) at (0.3*\d,1); 
			\coordinate (-xinputflow) at (0.5*\d,1); 
			
			\coordinate (-hinput) at (0,1); 
			\coordinate (-hinputflow) at (0,.9); 
			
			\coordinate (-houtput) at (1.1,1); 
			
			\coordinate (r)  at bl+(1*\d,0);
			\coordinate (rgap)at bl+(2*\d,0);
			\coordinate (u)  at bl+(3*\d,0);
			\coordinate (c)  at bl+(4*\d,0);
			
			\coordinate (out) at bl+(5*\d,0);
			
			\coordinate (abovegates) at (0,3.5*\h);
			\coordinate (gates) at (0,2.5*\h);
			\coordinate (belowgates) at ($(gates)!0.5!(bl)$);

			

			\node[gate](rgate) at ($ (gates -| r) $){$\VResetGate_t$};
			\node[gate](ugate) at ($ (gates -| u) $){$\VUpdateGate_t$};
			\node[gate](cgate) at ($ (gates -| c) $){$\tilde{\VHidden}_t$};
			
			\node[operator](cadd) at ($ (cgate |- bl) $) {$+$};
			\node[operator](cmult) at ($ (cgate |- belowgates) $) {$\odot$};
			
			\node[operator](rmult) at ($ (rgap |- belowgates) $) {$\odot$};
			
			\node[operator](umult) at ($ (u |- bl) $) {$\odot$};
			\draw[endflow] (ugate) -- (umult); 
			
			\draw[endflow] (-hinput)++(0,-.1) |- ($ (bl -| rgate) $) -| (rmult); 
			\draw[endflow] (rgate) |- (rmult);
			
			\draw[endflow] (cgate) -- (cmult);
			\draw[endflow] (ugate) |- (cmult);
			\draw[endflow] (cmult) -- (cadd);

			\draw[endflow] (-xinputflow) -| (rgate); 
			\draw[endflow] (-xinputflow) -| ($ (rgate |- abovegates) $) -| (ugate); 
			
			\draw[flow, draw=boxcolor,double=black,double distance=\pgflinewidth,ultra thick] (rmult) |- ($ (ugate |- tl) $);

			\draw[flow] (-xinputflow) -| (cgate); 
			\draw[flow] (-hinput)++(-0.1, 0) -- (-xinputflow); 
			
			\draw[flow] (-hinputflow) |- (umult) -- (cadd) -| ($(br)!0.5!(tr)$) |- (-houtput);

			
		\end{scope}
	}
}

\tikzset{pic shift/.store in=\shiftcoord,
	pic shift={(0,0)},
	concat/.pic = {
		\node[](a) at (0, .5){$\V{a}$};
		\node[](b) at (0, -.5){$\V{b}$};
		\node[](out) at (1, 0){$\concat{ \V{a} }{ \V{b} }$};
		
		\draw[endflow] (a) |- (out);
		\draw[endflow] (b) |- (out);
	}
}

\tikzset{pic shift/.store in=\shiftcoord,
	pic shift={(0,0)},
	copy/.pic = {
		\node[](ain) at (0, 0){$\V{a}$};
		\node[](aout1) at (.5, .5){$\V{a}$};
		\node[](aout2) at (.5, -.5){$\V{a}$};
		\draw[endflow] (ain) -| (aout1);
		\draw[endflow] (ain) -| (aout2);
	}
}

\tikzset{pic shift/.store in=\shiftcoord,
	pic shift={(0,0)},
	fgate/.pic = {
		\begin{scope}[shift={\shiftcoord},xscale=1,yscale=1]
			
			\node[dummy] (tl_a) at (0,0){}; 
			\node[dummy] (br_a) at (1,1){}; 
			
			\node[fit=(br_a) (tr_a),gate,inner sep=0] (-C) {};
			
			\node[draw] (conv) at (0.5,0){$conv$}; 
			\node[draw] (bn) at (0.5,.5){$bn$}; 
			\node[draw] (sigmoid) at (0.5,1){$\sigma$}; 
				
		\end{scope}
	}
}

\begin{tikzpicture}[scale=1, node distance=2em]





\draw pic (LSTM) at (0,0) {lstm};
\node[above=of LSTMtl](xt){$\VInput_{t}$};
\draw[rounded corners] (xt) |- (LSTM-input);
\node[left=of LSTMtl](htminus1){$\VHidden_{t-1}$};
\draw[endflow] (htminus1) -- (LSTM-input);
\node[right=of LSTMbr](ct){$\VCellState_{t}$};
\draw[endflow] (LSTM-coutput)--(ct);
\node[left=of LSTMbl](ctminus1){$\VCellState_{t-1}$};
\draw[endflow] (ctminus1)--(LSTMfmult);
\node[right=of LSTMtr](ht){$\VHidden_{t}$};
\draw[endflow] (LSTM-houtput)--(ht);

\draw pic (GRU) at (6.5,0) {gru};
\node[left=of GRU-hinput](gru_htminus1){$\VHidden_{t-1}$};
\draw[rounded corners] (gru_htminus1) -| (GRU-hinputflow);
\node[above=of GRU-xinput](gru_xt){$\VInput_{t}$};

\draw[rounded corners] (gru_xt) |- (GRU-xinputflow);

\node[right=of GRU-houtput](gru_ht){$\VHidden_{t}$};
\draw[rounded corners] (GRU-houtput)--(gru_ht);

\draw pic (concat_image) at (11,.5) {concat};
\node at (11.5,-.5) {\small concat};
\draw pic (copy_image) at (13,.5) {copy};
\node at (13.5,-.5) {\small copy};

%

\end{tikzpicture}
  \caption{%
    Schematic illustration of \glsfirst{lstm} and \glsfirst{gru} cells analog to the cell definitions in \cref{tab:rnn}.
    The cell output $\VHidden_t$ is calculated via internal gates and based on the current input $\VInput_t$ combined with prior context information $\VHidden_{t-1}$, $\VCellState_{t-1}$.
    \Gls{lstm} cells are designed to separately accommodate \emph{long-term} context in the internal cell state $\VCellState_{t-1}$, from \emph{short-term} context $\VHidden_{t-1}$.
    \Gls{gru} cells combine all context information in a single, but more sophisticated output $\VHidden_{t-1}$.
    \glsreset{lstm}
    \glsreset{gru}
    }
  \label{fig:lstmcell}
\end{figure}
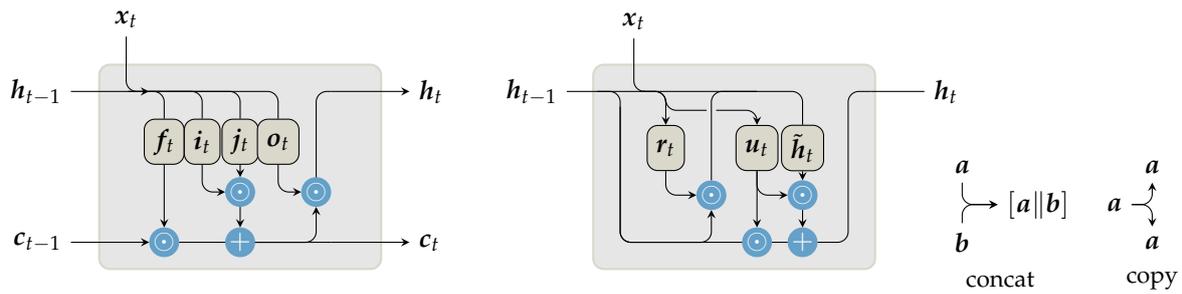

\Acrlongpl{ann} approximate a function $\hat{\V{y}} = f(\V{x};\V{W})$ of outputs $\hat{\V{y}}$ (e.g.,~{class~labels}) from {input data} $\V{x}$ given a large set of {weights} $\V{W}$. 
This approximation is commonly referred to as the {inference} phase.
These networks are typically composed of multiple cascaded layers with {hidden vectors} $\VHidden$ as intermediate layer outputs.
Analogous to the biological neural cortex, single~elements in these vectors are often referred to as {neurons}.
The quality of the approximation $\hat{\V{y}}$ with respect to ground truth $\V{y}$ is determined by the {loss function} $L(\hat{\V{y}},\V{y})$.
Based on this function, gradients are {back-propagated} through the \gls{ann} and adjust network weights $\V{W}$ at each {training} step.

Popular \acrlongpl{ffn} often utilize {convolutional} or {fully-connected} layers at which the input data are propagated through the network once.
This is realized by an affine transformation (fully-connected) $\VHidden=\sigma(\V{W}\VInput)$ or a convolution $\VHidden=\sigma(\conv{\V{W}}{\VInput})$ followed by an element-wise, non-linear transformation $\sigma: \mathbb{R} \mapsto \mathbb{R}$.

However, domains like translation~\citep{Bahdanau14}, text summarization~\citep{Rush15,Shen16, Nallapati16} or speech recognition~\citep{Sutskever14, Chorowski15} formulate input vectors naturally as a sequence of observations $\VInput=\{\VInput_{0},\dots,\VInput_T\}$.
In these domains, individual samples are generally less expressive, and the overall model performance is based largely on contextual information.

Sequential data are commonly processed with \acrfull{rnn} layers, in which the hidden layer output $\VHidden_t$ is determined at time $t$ by current input $\VInput_t$ in combination with the previous output $\VHidden_{t-1}$.
In theory, the iterative update of $\VHidden_t$ enables \glspl{rnn} to simulate arbitrary procedures~\citep{graves14}, since these networks are {Turing complete}~\citep{siegelmann1996}.
The standard \gls{rnn} variant performs the update step \mbox{$\VHidden_t=\sigma(\V{W}\tilde{\VInput})$} by an affine transformation of the concatenated vector $\tilde{\VInput} = \concat{\VInput_t}{\VHidden_{t-1}}$ followed by a non-linearity $\sigma$.
Consequently, the internal weight matrix is multiplied at each iteration step, which essentially raises it to a high power~\cite{Jozefowicz15}.
At gradient back-propagation, this iterative matrix multiplication leads to {vanishing} and {exploding gradients}~\citep{Hochreiter2001, Bengio1994}.
While exploding gradients can be avoided with {gradient clipping}, vanishing gradients impede the extraction of long-term feature relationships.
This issue has been addressed by~\citet{Hochreiter97:LST}, who introduced additional gates and an internal {state} vector $\VCellState_t$ in \gls{lstm} cells to control the gradient propagation through time and to enable {long-term} learning, respectively.
Analogous to standard \glspl{rnn}, the {output gate} $\VOutputGate_t$ balances the influence of the previous cell output $\VHidden_{t-1}$ and the current input $\VInput_t$.
At \glspl{lstm}, the cell output $\VHidden_{t}$ is further augmented by an internal state vector $\VCellState_t$, which is designed to contain long-term information.
To avoid the aforementioned vanishing gradients, reading and writing to the cell state is controlled by three additional gates.
The {forget gate} $\VForgetGate_t$ decreases previously-stored information by element-wise multiplication $\VCellState_{t-1} \odot \VForgetGate_t$.
New information is added by the product of {input gate} $\VInputGate_t$ and {modulation gate} $\VModulationGate_t$.
Illustrations of the internal calculation can be seen in \cref{fig:lstmcell}, and~the mathematical relations are shown in \cref{tab:rnn}.
Besides \glspl{lstm}, \glspl{gru}~\citep{Cho2014} have gained increasing popularity, as these cells achieve similar accuracies to \glspl{lstm} with fewer trainable parameters.
Instead of separate vectors for long- and short-term memory, \glspl{gru} formulate a single, but more sophisticated, output vector.

To account for the more complicated design, recurrent layers are conventionally referred to as a collection of {cells} with a single cell representing the set of elements at one vector-index.

\begin{figure}
 	\subfigure[%
 	Network structure employed in previous work \citep{Russwurm17:TVM}.
 	]{
 		
 		\hspace{-5em}
 		\raisebox{1ex}{
 		\input{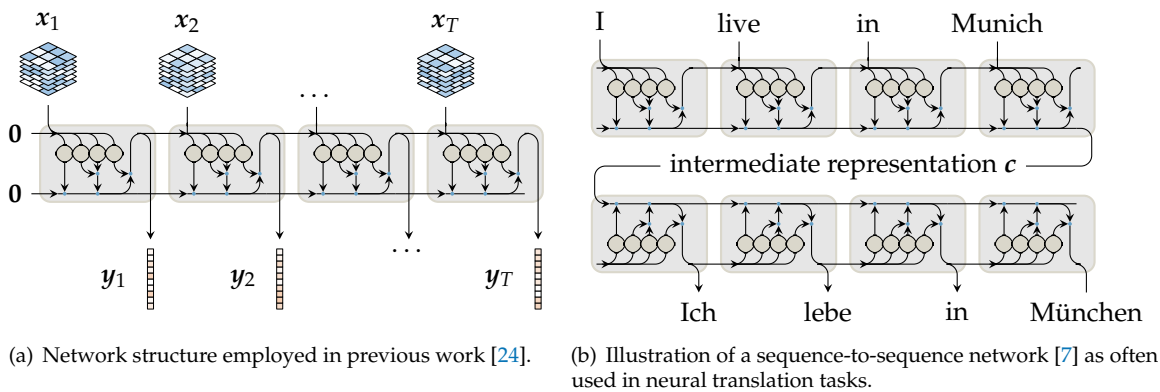}\label{fig:fieldRNN_encoder}
	 	}
 	}
  \subfigure[%
  Illustration of a sequence-to-sequence network \cite{Sutskever14} as often used in neural translation tasks.
  ]{
	\hspace{-3em}
  	\tikzsetnextfilename{seq2seq}

\tikzstyle{operator} = [draw, circle, fill=tumbluemedium, draw=tumbluemedium, inner sep=0, text=white]
\tikzstyle{gate} = [fill=tumivory,draw,rounded corners]

\tikzstyle{textnode} = [scale=1]

\tikzstyle{dummy} = [inner sep=0]
\tikzstyle{flow} = [rounded corners]
\tikzstyle{endflow} = [-stealth,flow]
\tikzstyle{beginflow} = [stealth-,flow]

\tikzstyle{bigbox} = [rectangle, draw=tumivory, thick, fill=tumgraylight, rounded corners, 
inner sep=.5ex]

\tikzset{pic shift/.store in=\shiftcoord,
	pic shift={(0,0)},
	pics/seqlstmencoder/.style={
		code={
		\begin{scope}[shift={\shiftcoord},xscale=1.3,yscale=.8]
			
			\node[dummy] (bl) at (0,0){}; 
			\node[dummy] (tr) at (1,1){}; 
			
			\node[dummy] (br) at ($ (bl -| tr) $){}; 
			\node[dummy] (tl) at ($ (bl |- tr) $){}; 
			
			\node[fit=(bl) (tr),bigbox] (-C) {};
			
			\coordinate (-input) at (0.1,1); 
			
			\coordinate (-coutput) at (0.9,0); 
			\coordinate (-cinput) at (0.1,0); 
			\coordinate (-houtput) at (0.9,1); 
			
			\def\d{1/6}
			
			\def\h{1/3}
			
			\coordinate (f)  at bl+(1*\d,0);
			\coordinate (i)  at bl+(2*\d,0);
			\coordinate (j)  at bl+(3*\d,0);
			\coordinate (o)  at bl+(4*\d,0);
			\coordinate (out) at bl+(5*\d,0);
			
			\coordinate (gates) at (0,2*\h);
			
			
			
			\node[gate](fgate) at ($ (gates -| f) $){};
			\node[gate](igate) at ($ (gates -| i) $){};
			\node[gate](jgate) at ($ (gates -| j) $){};
			\node[gate](ogate) at ($ (gates -| o) $){};
			
%
			\node[operator](fmult) at ($ (bl -| fgate) $) {};
			\draw[endflow] (-input) -| (fgate) -- (fmult); 
			
			\node[operator](jmult) at ([shift={(0,-1*\h)}]jgate) {};
			\node[operator](cadd) at ($ (bl -| jgate) $) {};
			\draw[endflow] (-input) -| (jgate) -- (jmult);
			\draw[endflow] (jmult) -- (cadd); 			

			\draw[endflow] (-input) -| (igate) |- (jmult); 
%
			\node[operator](outtanh) at ([shift={(0,1*\h)}]out) {};
%
			\draw[endflow] (tl) -| (ogate) |- (outtanh);
			\draw[flow] (outtanh) |- (-houtput);
%
			\draw[endflow] (cadd) -| (outtanh);
			\draw[flow] (-cinput) -- (fmult) -- (cadd) -- (-coutput);
%
			
			
		\end{scope}
		}
	}
}
\tikzset{pic shift/.store in=\shiftcoord,
	pic shift={(0,0)},
	pics/seqlstmdecoder/.style={
		code={
			\begin{scope}[shift={\shiftcoord},xscale=1.3,yscale=-.8]
				
				\node[dummy] (bl) at (0,0){}; 
				\node[dummy] (tr) at (1,1){}; 
				
				\node[dummy] (br) at ($ (bl -| tr) $){}; 
				\node[dummy] (tl) at ($ (bl |- tr) $){}; 
				
				\node[fit=(bl) (tr),bigbox] (-C) {};
				
				\coordinate (-input) at (0.1,1); 
				
				\coordinate (-coutput) at (0.9,0); 
				\coordinate (-cinput) at (0.1,0); 
				\coordinate (-houtput) at (0.9,1); 
				
				\def\d{1/6}
				
				\def\h{1/3}
				
				\coordinate (f)  at bl+(1*\d,0);
				\coordinate (i)  at bl+(2*\d,0);
				\coordinate (j)  at bl+(3*\d,0);
				\coordinate (o)  at bl+(4*\d,0);
				\coordinate (out) at bl+(5*\d,0);
				
				\coordinate (gates) at (0,2*\h);
				
				
				
				\node[gate](fgate) at ($ (gates -| f) $){};
				\node[gate](igate) at ($ (gates -| i) $){};
				\node[gate](jgate) at ($ (gates -| j) $){};
				\node[gate](ogate) at ($ (gates -| o) $){};
				
				%
				\node[operator](fmult) at ($ (bl -| fgate) $) {};
				\draw[endflow] (-input) -| (fgate) -- (fmult); 
				
				\node[operator](jmult) at ([shift={(0,-1*\h)}]jgate) {};
				\node[operator](cadd) at ($ (bl -| jgate) $) {};
				\draw[endflow] (-input) -| (jgate) -- (jmult);
				\draw[endflow] (jmult) -- (cadd); 			
				
				\draw[endflow] (-input) -| (igate) |- (jmult); 
				%
				\node[operator](outtanh) at ([shift={(0,1*\h)}]out) {};
				%
				\draw[endflow] (tl) -| (ogate) |- (outtanh);
				\draw[flow] (outtanh) |- (-houtput);
				%
				\draw[endflow] (cadd) -| (outtanh);
				\draw[flow] (-cinput) -- (fmult) -- (cadd) -- (-coutput);
				%
				
				
			\end{scope}
		}
	}
}

\begin{tikzpicture}[scale=1, node distance=1em]



\def\d{1.7}
\def\encoderheight{.5}
\def\decoderheight{-.5}

\draw pic (enc1) at (\d,\encoderheight) {seqlstmencoder};
\node[textnode,above=of enc1tl](xenc1){I};

\draw pic (enc2) at (2*\d,\encoderheight) {seqlstmencoder};
\node[textnode,above=of enc2-input](xenc2){live};

\draw pic (enc3) at (3*\d,\encoderheight) {seqlstmencoder};
\node[textnode,above=of enc3-input](xenc3){in};

\draw pic (enc4) at (4*\d,\encoderheight) {seqlstmencoder};
\node[textnode,above=of enc4-input](xenc4){Munich};

\draw pic (dec1) at (1*\d,\decoderheight) {seqlstmdecoder};
\node[textnode,below=of dec1tr](ydec1){Ich};

\draw pic (dec2) at (2*\d,\decoderheight) {seqlstmdecoder};
\node[textnode,below=of dec2tr](ydec2){lebe};

\draw pic (dec3) at (3*\d,\decoderheight) {seqlstmdecoder};
\node[textnode,below=of dec3tr](ydec3){in};

\draw pic (dec4) at (4*\d,\decoderheight) {seqlstmdecoder};
\node[textnode,below=of dec4tr](ydec4){München};

\node[textnode,anchor=center](state) at ($(enc4-coutput)!0.5!(dec1-cinput)$){intermediate representation $\VCellState$};

\node[left=.5emof enc1-input](enczerostateh){};
\node[left=.5emof enc1-cinput](enczerostatec){};
\node[left=.5emof dec1-input](deczerostateh){};

\draw[endflow] (enczerostateh) -- (enc1-input);
\draw[endflow] (enczerostatec) -- (enc1-cinput);
\draw[endflow] (deczerostateh) -- (dec1-input);

\draw[flow] (enc4-coutput) -- ++(.2,0) |- (state);
\draw[beginflow] (dec1-cinput) -- ++(-.2,0) |- (state);

\draw[flow] (xenc1) |- (enc1-input);
\draw[flow] (xenc2) |- (enc2-input);
\draw[flow] (xenc3) |- (enc3-input);
\draw[flow] (xenc4) |- (enc4-input);

\draw[beginflow] (ydec1) |- (dec1-houtput);
\draw[beginflow] (ydec2) |- (dec2-houtput);
\draw[beginflow] (ydec3) |- (dec3-houtput);
\draw[flow] (ydec4) |- (dec4-houtput);

\draw[endflow] (enc1-houtput) -- (enc2-input);
\draw[endflow] (enc2-houtput) -- (enc3-input);
\draw[endflow] (enc3-houtput) -- (enc4-input);

\draw[endflow] (dec1-houtput) -- (dec2-input);
\draw[endflow] (dec2-houtput) -- (dec3-input);
\draw[endflow] (dec3-houtput) -- (dec4-input);

\draw[endflow] (enc1-coutput) -- (enc2-cinput);
\draw[endflow] (enc2-coutput) -- (enc3-cinput);
\draw[endflow] (enc3-coutput) -- (enc4-cinput);

\draw[endflow] (dec1-coutput) -- (dec2-cinput);
\draw[endflow] (dec2-coutput) -- (dec3-cinput);
\draw[endflow] (dec3-coutput) -- (dec4-cinput);


\end{tikzpicture}\label{fig:seq2seq}
  }%
  
  \caption{%
  	Illustrations of recurrent network architectures which inspired this work. 
  	The network of previous work \citep{Russwurm17:TVM} shown in \cref{fig:fieldRNN_encoder} creates a prediction $\V{y}_t$ at each observation $t$ based on spectral input information $\VInput_t$ and the previous context $\VHidden_{t-1}$, $\VCellState_{t-1}$.
  	Sequence-to-sequence networks, as shown in \cref{fig:seq2seq}, aggregate sequential information to an intermediate state $\VCellState_T$ which is a representation of the entire series.
  }

\end{figure}

%

The common output of recurrent layers provides a {many-to-many} relation by generating an output vector at each observation $\VHidden_t$ given previous context $\VHidden_{t-1}$ and $\VCellState_{t-1}$, as shown in Figure~\ref{fig2}a.
However, encoding information of the entire sequence in a {many-to-one} relation is favored in many applications.
Following this idea, {sequence-to-sequence} learning, illustrated in Figure~\ref{fig2}b, has popularized the use of the cell state vector $\VCellState_T$ at the last-processed observation $T$ as a representation of the entire input sequence.
These encoding-decoding networks transform an input sequence of varying length to an intermediate state representation $\VCellState$ of fixed size.
Subsequently, the decoder generates a varying length output sequence from this intermediate representation.
Further developments in this domain include {attention schemes}.
These provide additional intermediate connections between encoder and decoder layers, which are beneficial for translations of longer sequences~\citep{Bahdanau14}.

In many sequential applications, the common input form is $\VInput_t \in \mathbb{R}^{d}$ with a given {depth} $d$.
The~output vectors $\VHidden_t \in \mathbb{R}^{r} $ are computed by matrix multiplication with internal weights $\MWeight \in \mathbb{R}^{(r+d) \times r} $ and $r$ recurrent cells.
However, other fields, such as image processing, commonly handle raster data $\VInput_t \in \Rin{h}{w}{d}$ of specific {width} $w$, {height} $h$ and spectral {depth} $d$.
To account for neighborhood relationships and to circumvent the increasing complexity, convolutional variants of \glspl{lstm}~\cite{Shi15} and \glspl{gru} have been introduced.
\newtext{
	These variants convolve the input tensors with weights $W \in \Rinfour{k}{k}{(r+d)}{r}$ augmented by the convolutional kernel size $k$, which is a hyper-parameter determining the perceptive field.
}




\subsection{Prior Work}
\label{sec:prevwork}

\newtext{
	Given recurrent networks as popular architectures for sequential data processing, we~experimented with recurrent layers for multi-temporal vegetation classification prior to this work~\cite{Russwurm17:TVM}.
	In the conducted experiments, we used a network architecture similar to the illustration in Figure~\ref{fig2}a.
	Following the input dimensions of standard recurrent layers, an input sequence $\VInput \in \{\VInput_0,$\ldots$,\VInput_T \}$ of observations $\VInput_t \in \mathbb{R}^d$ was introduced to the network.
	Based on contextual information from previous observations, a classification for each observation $\V{y}_t$ was produced.
	We~evaluated the effect of this information gain by comparing the recurrent network with \glspl{cnn} and a \gls{svm}.
	Standard \glspl{rnn} and \glspl{lstm} outperformed their non-sequential \glspl{svm} and \glspl{cnn} counterparts.
	Further, we observed an increase in accuracy at sequentially later observations, which were classified with more context information available.
	Overall, we concluded that recurrent network architectures are well suited for the extraction of temporal features from multi-temporal \gls{eo} imagery, which is consistent with other recent findings~\citep{Jia2017b,Sharma17,Braakmann-Folgmann2017}.
	
	However, the experimental setup introduced some limitations regarding applicability in real-world scenarios.
	We followed the standard formulation of recurrent networks, which process a $d$-dimensional input vector.
	This vector included the concatenated \gls{boa} reflectances of nine pixels neighboring one {point-of-interest}.
	The point-wise classification was sufficient for quantitative accuracy evaluation, but could not produce areal classification maps.
	Since a class prediction was performed on every observation, we introduced additional {covered} classes for cloudy pixels at single images.
	These were derived from the scene classification of the \brand{sen2cor}
	atmospheric correction algorithm, which required additional preprocessing.
	A single representative classification for the entire time-series would have required additional post-processing to further aggregate the predicted labels for each observation.
	Finally, the mono-directional iterative processing introduced a bias towards last observations.
	With more contextual information available, later observation showed better classification accuracies compared to observations earlier in the sequence.
	

}

\subsection{This Approach}
\label{sec:approach}

\newtext{
	To address the limitations of previous work, we redesigned and streamlined the network structure and processing pipeline.
	Inspired by sequence-to-sequence structures described in \cref{sec:methodology}, the~proposed network aggregates the information encoded in the cell state $\VCellState_t$ within the recurrent cell.
	Since one class prediction for the entire temporal series is produced, atmospheric perturbations can be treated as temporal noise.
	Hence, explicitly introduced cloud-related labels are not required, which~alleviates the need for prior cloud classification.
	Without the need for prior scene classification to obtain these classes, the performance on atmospherically uncorrected \gls{toa} reflectance data can be evaluated.
	We further implemented convolutional recurrent cell variants, as~formulated in \cref{tab:rnn}, to process input tensors $\VInput_t$ of given height $h$, width $w$ and depth $d$.
	Hence, the proposed network produces areal prediction maps as shown in the qualitative results \cref{sec:qualitative}.
	Finally, we introduce the input sequence in a bidirectional manner to eliminate any bias towards the later elements in the observation sequence.
}

Overall, we employ a bidirectional sequential encoder for the task of multi-temporal land cover classification.
As Earth observation data are gathered in a periodic manner, many observations of the same area at consecutive times are available, which may contribute to the classification decision.
Inspired by sequence-to-sequence models, the proposed model encodes this sequence of images into a fixed-length representation.
Compared to previous work, this is an elegant way to condense the available temporal dimension without further post-processing.
A classification map for each class is derived from this sequence representation.
Many optical observations are covered by clouds, and prior cloud classification is often required as additional preprocessing step.
As clouds do not contribute to the classification decision, these observations can be treated as temporal noise and may be potentially ignored by this encoding scheme.
In \cref{sec:activations}, we investigate this by visualizing internal activation states on cloudy and non-cloudy observations.

\Cref{fig:network} presents the proposed network structure schematically.
The input image sequence $\VInput = \{\VInput_t,\dots,\VInput_T\}$ of observations $\VInput \in \Rin{h}{w}{d}$ is passed to gated recurrent layers at each observation time $t$.
The index $T$ denotes the maximum length of the sequence and $d$ the input feature depth.
In practice, sequence lengths are often shorter than $T$, as the availability of satellite acquisitions is variable over larger scales.
If less than $T$ observations are present, sequence elements are padded with a constant value and are subsequently ignored at the iterative encoding steps.
To eliminate bias towards the last observations in the sequence, the data are passed to the encoder in both {sequential (seq)} and {reversed (rev)} order.
Network weights are shared between both passes.
The initial cell states $\VCellState_0^\text{seq},\VCellState_T^\text{rev} \in \Rin{h}{w}{r}$ and output $\VHidden_0^\text{seq},\VHidden_T^\text{rev} \in \Rin{h}{w}{r}$ are initialized with zeros.
The concatenated final states $\VCellState_T = \concat{\VCellState_T^\text{seq}}{\VCellState_0^\text{inv}}$ are the representation of the entire sequence and are passed to a convolutional layer for classification.
A second convolutional classification layer projects the sequence representation $\VCellState_T$ to {softmax-normalized} activation maps $\V{\hat{y}}$ for $n$ classes: $\VCellState_T \in \Rin{h}{w}{2r} \mapsto \V{\hat{y}} \in \Rin{h}{w}{n}$.
This layer is composed of a convolution with a kernel size of $\kclass$, followed by batch normalization and a {\gls{relu}}~\cite{Hahnloser2000} or {leaky \gls{relu}}~\cite{Maas13} non-linear activation function. 
{ 
	At each training step, the cross-entropy loss
	\begin{align}
	H(\hat{\V{y}},\V{y}) &= -\sum_{i} y_i log(\hat{y}_i)
	\end{align}
	between the predicted activations $\hat{\V{y}}$ and an one-hot representation of the ground truth labels $\V{y}$ evaluates the prediction quality.

Tunable hyper-parameters are the number of recurrent cells $r$ and the sizes of the convolutional kernel $\krnn$ and the classification kernel $\kclass$. 

\begin{figure}
  \input{images/network.tikz}
  \centering
  \caption{%
    Schematicial illustration of our proposed bidirectional sequential encoder network.
    The input sequence $\VInput\in\{\VInput_0,\dots,\VInput_T\}$ of observations $\VInput_t \in \Rin{h}{w}{d}$ is encoded to a representation $\VCellState_T = \concat{\VCellState_T^\text{seq}}{\VCellState_0^\text{inv}}$.
    The observations are passed in sequence (seq) and reversed (rev) order to the encoder to eliminate bias towards recent observations.
    The concatenated representation of both passes $\VCellState_T$ is then projected to softmax-normalized feature maps for each class using a convolutional layer.
    }
  \label{fig:network}
\end{figure}

\section{Dataset}
\label{sec:data}

\begin{figure}
  \centering
  \includegraphics{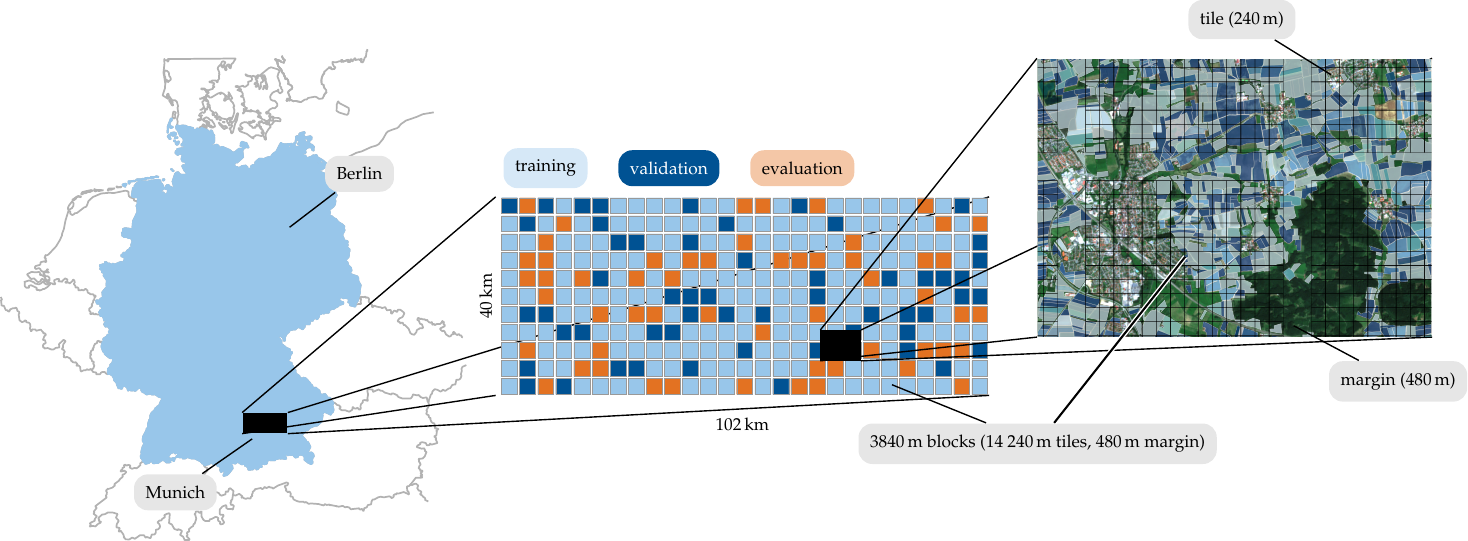}
  \caption{
  \Glsfirst{aoi} north of Munich containing \SI{430}{\kilo\hectare} and \SI{137}{\kilo\nothing} field parcels.
  The \gls{aoi} is further tiled at multiple scales into datasets for training, validation and evaluation and footprints of individual samples.
  }
  \label{fig:aoi}
\end{figure}

For the evaluation of our approach, we defined a large \acrfull{aoi} of \SI{102}{\km} $\times$ \SI{42}{\km} north of Munich, Germany.
An overview of the \gls{aoi} at multiple scales is shown in \cref{fig:aoi}.
The~\gls{aoi} was further subdivided into squared blocks of \SI{3.84}{\km} $\times$ \SI{3.84}{\km} (multiples of \SI{240}{\m} and \SI{480}{\m}) to ensure dataset independence while maintaining similar class distributions.
These blocks were then randomly assigned to partitions for network training, hyper-parameter validation and model evaluation in a ratio of 4:1:1 similar to previous work~\citep{Russwurm17:TVM}.
The spatial extent of single samples $\VInput$ is determined by tile-grids of \SI{240}{\meter} and \SI{480}{\meter}.
We bilinearly interpolated the \SI{20}{\meter} and \SI{60}{\meter} \gls{s2} bands to \SI{10}{\meter} \gls{gsd} to harmonize the raster data dimensions.
To provide additional temporal meta information, the {year} and {day-of-year} of the individual observations were added as matrices to the input tensor.
Hence, the input feature depth $d=15$ is composed of four \SI{10}{\meter} (\band{B4}, \band{B3}, \band{B2}, \band{B8}), six \SI{20}{\meter} (\band{B5}, \band{B6}, \band{B7}, \band{B8A}) and three \SI{60}{\meter} (\band{B1}, \band{B11}, \band{B12}) bands combined with {year} and {day-of-year}.

With ground truth labels of two growing seasons 2016 and 2017 available, we gathered 274 (108~in 2016; 166 in 2017) \acrlong{s2} products at 98 (46 in 2017; 52 in 2017) observation dates between 3~January~2016 and 15~November~2017.
The obtained time series represents all available \gls{s2} products labeled with cloud coverage less than 80\%.
\newtext{
	In some \gls{s2} images, we noticed a spatial offset in the scale of one pixel.
	However, we did not perform additional georeferencing and treated the spatial offset as data-inherent observation noise.
	Overall, we relied on the geometrical and spectral reference as provided by the \brand{Copernicus} ground segment.
}

Ground truth information was provided by the \gls{stmelf} in the form of geometry and semantic labels of \SI{137}{\kilo\nothing} field parcels.
The crop-type is reported by farmers to the ministry as mandated by the European crop subsidy program.
We selected and aggregated 17 crop-classes from approximately 200 distinct field labels, occurring at least 400 times in the AOI.
With modern agriculture, centered on a few predominant crops, the distribution of classes is not uniform, as can be observed from Figure~\ref{fig:combinedaoi}a.
This non-uniform class distribution is generally not optimal for the classification evaluation as it skews the overall accuracy metric towards classes of high frequency.
Hence, we additionally calculated kappa metrics~\citep{cohen1960} for the quantitative evaluation in \cref{sec:quantitative} to compensate for unbalanced distributions.



\begin{figure}[b]
	\subfigure[Non-uniform distribution of field classes in the \gls{aoi}]{\usepgfplotslibrary{groupplots}

\tikzsetnextfilename{classhistogram}
\begin{tikzpicture}

\def\data{images/classhist.csv}
  \pgfplotsset{ every non boxed x axis/.append style={x axis line style=-},
  	every non boxed y axis/.append style={y axis line style=-}}

  \pgfplotsset{every axis/.append style={ybar=1pt, bar width=3pt, ymajorgrids}}
  \pgfplotsset{every axis label/.append style={font=\footnotesize},tick pos=left,ylabel near ticks}
  \pgfplotsset{every tick label/.append style={font=\footnotesize}}
  \pgfplotsset{every x tick label/.append style={rotate=35,anchor=north east,font=\footnotesize}}
  \tikzstyle{caption}=[font=\footnotesize, fill=tumwhite, fill opacity=.5, text opacity=1]

  \begin{groupplot}[
    group style={
      group size=1 by 1,
      xlabels at=edge bottom,
      xticklabels at=edge bottom,
      ylabels at=edge left,
      yticklabels at=edge left,
      vertical sep=2pt,
      horizontal sep=2pt
    },
    width=.5\textwidth,
    height=3cm,
    ymode=log,
    log origin=infty,
    log ticks with fixed point,
    scaled y ticks=false,
    axis lines=left,
    xlabel={crop classes},
    xlabel style={yshift=-7mm},
    xmin=1,
    xmax=15,
    ymin=0,
    ymax=32000,
    ytick={100,300,1000,3000,10000,30000},
    enlarge x limits,
    enlarge y limits=.1,
    restrict y to domain=0:30000,
    xtick=data,
    major grid style={draw=tumgraylight},
    xticklabels={
      \classname{maize},
      \classname{wheat},
      \classname{meadow},
      \classname{winter barley},
      \classname{potatoe},
      \classname{rapeseed},
      \classname{summer barley},
      \classname{hop},
      \classname{triticale},
      \classname{oat},
      \classname{rye},
      \classname{sugar beet},
      \classname{spelt},
      \classname{asparagus},
      \classname{beans},
      \classname{peas},
      \classname{soybeans}
    },
    legend style={draw=none, fill=tumgraylight, fill opacity=1, text opacity=1, font=\footnotesize, rounded corners=1pt, inner sep=2pt,at={(1,.9)}},
    legend columns=-1,
  ]

    \nextgroupplot[ylabel=field parcels]
    
    \addplot[
      draw=none,
      fill=tumblue,rounded corners=.5pt
    ] table [col sep=comma, x=id, y=count16] {\data};
    \addlegendentry{2016}
   
    \addplot[
    draw=none,
    fill=tumgray,rounded corners=.5pt
    ] table [col sep=comma, x=id, y=count17] {\data};
    \addlegendentry{2017}

  \end{groupplot}
\end{tikzpicture}\label{fig:classhist}}
	\subfigure[Acquired \acrfull{s2} observations of the twin satellites S2A and S2B]{\raisebox{6mm}{

\tikzsetnextfilename{observations}
\begin{tikzpicture}

\pgfplotsset{every tick label/.append style={font=\footnotesize}}
\pgfplotsset{every mark/.append style={fill=tumblue}}

    \begin{axis}[
    xmajorgrids=false,
    x axis line style={draw=none},
    width=.5\textwidth,
    height=3cm,
    date coordinates in=x,
    xticklabel={\month},
    date ZERO=2016-01-01, 
    xmin={2015-12-15}, 
    xmax={2017-12-30},
    ymin=-.5,
    ymax=1.5,
    mark size=1.25pt,
	xlabel style={yshift=-4ex},
    extra x ticks={2016-01-01,2017-01-01},
    extra x tick labels={2016,2017},
    extra x tick style={
    	grid=none,
    	tick label style={
	    		yshift=-3ex
    		}
    	},
    ytick={0,  1},
    yticklabels={\satellite{S2A},\satellite{S2B}},
    y tick style={draw=none, xshift=-5ex},
    ]
        \addplot [
        tumblue,
        only marks, 
        opacity=.5
        ]
        table[x=date,y=sat_id, col sep=comma]{images/observations.csv};
    \end{axis}
\end{tikzpicture}}\label{fig:observations}}
	
	\caption{%
		Information of the \acrlong{aoi} containing location, division schemes, class distributions and dates of acquired satellite imagery.  
	}
	\label{fig:combinedaoi}
\end{figure}
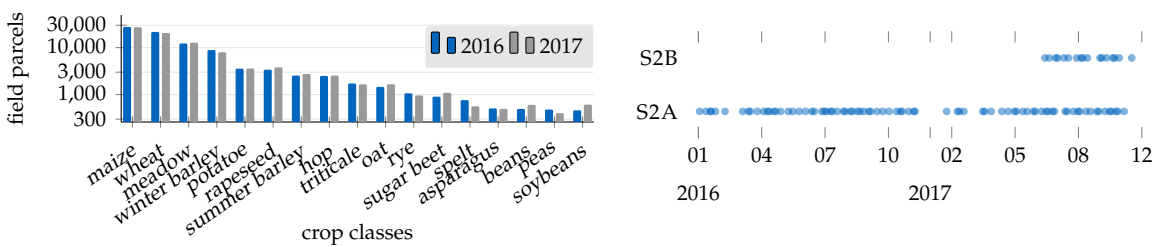

\section{Results}
\label{sec:results}

\newtext{
In this section, we first visualize internal state activations in \cref{sec:activations} to gain a visual understanding of the sequential encoding process.
Further findings on internal cloud masking are presented before the classification results on crop classes are quantitatively and qualitatively evaluated in \cref{sec:quantitative,sec:qualitative}.
}

\subsection{Internal Network Activations}
\label{sec:activations}

In \cref{sec:seqenc}, we gave an overview of the functionality of recurrent layers and discussed the property of \gls{lstm} state vectors $\VCellState_t \in \Rin{h}{w}{r}$ to encode sequential information over a series of observations.
The cell state is updated by internal gates $\VInputGate_t, \VModulationGate_t, \VForgetGate_t \in \Rin{h}{w}{r}$, which in turn are calculated based on previous cell output $\VHidden_{t-1}$ and cell state $\VCellState_{t-1}$ (see \cref{tab:rnn}).
To assess prior assumptions regarding cloud filtering and to visually assess the encoding process, we visualized internal \gls{lstm} cell tensors for a sequence of images and show representative activations of three cells in \cref{fig:activations}.
The~\gls{lstm} network, from which these activations are extracted, was trained on \SI{24}{\pixel} $\times$ \SI{24}{\pixel} tiles with $r=256$ recurrent cells and $\krnn=\kclass=3$ px.
Additionally, we inferred the network with tiles of height $h$ and width $w$ of \SI{48}{\pixel}.
Experiments with the input size of \SI{24}{\pixel} show similar results and are included in the Supplementary Material to this work.
In the first row, a $4\sigma$ band-normalized RGB image represents the input satellite image $\VInput_t \in \Rinfloor{h=48}{w=48}{d=15}$ at each time frame $t$.
The~next rows show the activations of input gate $\VInputGate_t^i$, modulation gate $ \VModulationGate_t^i $, forget gate $\VForgetGate_t^i$ and cell state $\VCellState_t^i$ at three selected recurrent cells, which are denoted by the raised index $i \in \{3,22,47\}$.
After iteratively processing the sequence, the final cell state $\VCellState_{T=36}$ is used to produce activations for each class, as~described in \cref{sec:approach}.

In the encoding process, the detail of structures at the cell state tensor increased gradually.
This~may be interpreted as additional information written to the cell state.
It further appeared that the structures visible at the cell states resembled shapes, which were present in cloud-free RGB images (e.g., $\VCellState_{t=15}^{(3)}$ or $\VCellState_{t=28}^{(22)}$).
Some cells (e.g., Cell 3 or Cell 22) changed their activations gradually over the span of multiple observations, while others (e.g., 48) changed more frequently.
Forget gate $\VForgetGate$ activations are element-wise multiplied with the previous cell state $\VCellState_{t-1}$ and range between zero and one.
Low~values in this gate numerically reduce the cell state, which can be potentially interpreted as a change of decision.
The input $\VInputGate$ and modulation gate $\VModulationGate$ control the degree of new information written to the cell state.
While the input gate is scaled between zero and one, the modulation gate $\VModulationGate \in [-1,1]$ determines the sign of change.
In general, we found the activity of a majority of cells (e.g., Cell 3 or Cell 22) difficult to associate with distinct events in the current input.
However, we assumed that classification-relevant features were expressed as a combination of cell activations similar to other neural network approaches.
Nevertheless, we could identify a proportionally small number of cells, in~which the shape of clouds visible in the image was projected on the internal state activations.
One~of these was cell $i=47$.
For cloudy observations, the input gate approached zero either over the entire tile (e.g., $t=\{10,18,19,36\}$) or over patches of cloudy pixels (e.g., $t=\{11,13,31,33\}$).
At some observation times (e.g., $t=\{13,31,32\}$), the modulation gate $\VModulationGate_t^{(47)}$ additionally changed the sign.

In a similar fashion,~\citet{Karpathy2015} evaluated cell activations for the task of text processing.
He~could associate a small number of cells with a set of distinct tasks, such as monitoring the lengths of a sentence or maintaining a state-flag for text inside and outside of brackets.

Summarizing this experiment, the majority of cells showed increasingly detailed structures when new information was provided in the input sequence.
It is likely that the grammar of crop-characteristic phenological changes was encoded in the network weights, and we suspect that a certain amount of these cells was sensitive to distinct events relevant for crop identification.
However, these events may be encoded in multiple cells and were difficult to visually interpret.
A small set of cells could be visually associated with individual cloud covers and may be used for internal cloud masking.
Based~on these findings, we are confident that our network has learned to internally filter clouds without explicitly introducing cloud-related labels.

\begin{landscape}
  \begin{figure}

    \newcounter{tcounter}
    
    
    \includegraphics{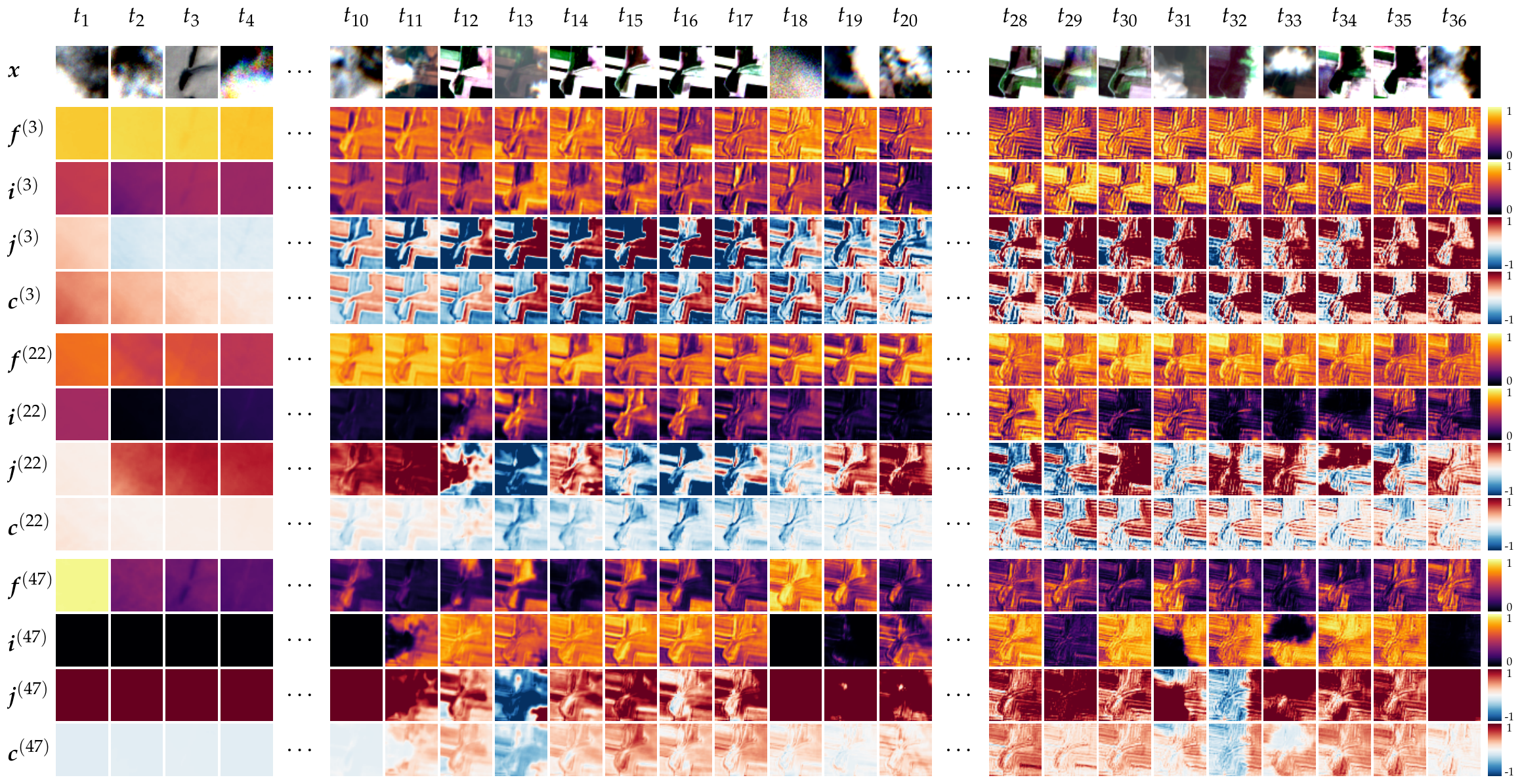}

    \caption{Internal \gls{lstm} cell activations of input gate $\VInputGate^{(i)}$, forget gate $\VForgetGate^{(i)}$, modulation gate $\VModulationGate^{(i)}$ and cell state $\VCellState^{(i)}$ at three (of $r=256$) selected cells $i\in\{3,22,47\}$ given the current input $\VInput_t$ over the sequence of observations $t=\{1,..,36\}$.
    The detail of features at the cell states increased gradually, which indicated the aggregation of information over the sequence.
    While most cells likely contribute to the classification decision, only some cells are visually interpretable with regard to the current input $\VInput_t$. 
    One visually interpretable cell $i=47$ has learned to identify cloud, as input and modulation gates show different activation patterns on cloudy and non-cloudy observations. 
    }
    \label{fig:activations}
  \end{figure}
\end{landscape}

\subsection{Quanititative Classificaton Evaluation}
\label{sec:quantitative}

\tabAccuracies

For the quantitative evaluation of our approach, we trained networks with bidirectional convolutional \gls{lstm} and \gls{gru} cells with $r\in\{128,256\}$ recurrent cells.
Kernel sizes of $\krnn = \kclass = 3$ were used for the evaluation since previous tests with larger kernel sizes showed similar accuracies.
For these initial experiments, we predominantly tested network variants with $r=128$ recurrent cells, as these networks could be trained within a reasonable time frame.
We decided to use networks with $r=256$ recurrent cells for the final accuracy evaluation, as we found that these variants achieved slightly better results in prior tests.
The convolutional \Gls{gru} and \Gls{lstm} networks were trained on a \brand{P100} GPU for 60 epochs (3.51 Mio \SI{24}{\pixel} $\times$ \SI{24}{\pixel} tiles seen) and took \SI{58}{\hour} and \SI{51}{\hour}, respectively.
However, reasonable accuracies were achieved within the first twelve hours, and further training increased the accuracies on validation data only marginally.
At each training step, a subset of 30 observations was randomly sampled from all available 46 (2016) and 52 (2017) observations to randomize the sequence while the sequential order was maintained.
For all our tests, the performance of \gls{lstm} and \gls{gru} networks was similar. 
The fewer weights of \gls{gru} cells, however, allowed using a slightly larger batch size of 32 samples compared to 28 samples of the \gls{lstm} variant.
This led to a seven-hour faster training compared to the \gls{lstm} variant.


For these reasons, we decided to report evaluation results of the \gls{gru} network in \cref{tab:accuracies}.
\newtext{
	Precision~and recall are common accuracy measures that normalize the sum of correctly-predicted samples with the total number of predicted and reference samples of a given class, respectively.
	These~measures are equivalent to user's and producer's accuracies and inverse to errors of commission and omission, which are popular metrics in the remote sensing community.
	We further calculated the $f$-measure as the harmonic average of precision and recall and the overall accuracy as the sum of correctly-classified samples normalized by the total number of samples.
	These metrics weight each sample equally.
	This introduces a bias towards frequent classes in the dataset, such as {maize} or {wheat}.
	To compensate for the non-uniform class distribution, we additionally report the conditional~\citep{Fung1988} and overall kappa~\cite{cohen1960} coefficients, which are normalized by the probability of a hypothetical correct classification by chance.
	The kappa coefficient $\kappa$ is a measure of agreement and typically ranges between $\kappa=0$ for no and $\kappa=1$ for complete agreement.
	\citet{McHugh2012} provides an interpretative table in which values $0.4 \leq \kappa < 0.6$ are considered `weak', values $0.6 \leq \kappa < 0.8$ `moderate', $0.8 \leq \kappa \leq 0.9$ considered `strong' and values beyond $0.9$ `almost perfect'.
}

The provided table of accuracies shows precision, recall, $f$-measure and the conditional kappa coefficient for each class over the two evaluated seasons.
Furthermore, overall accuracy and overall kappa coefficients indicate the quality for the classification and report good accuracies.
The~pixel-averaged achieved precision, recall and {\textit{f}}-score accuracies were consistent and ranged between 89.3\% and 89.9\%.
The kappa coefficients of $0.870$ and overall accuracies of 89.7\% and 89.5\% show similar consistency.
While these classification measures reported good performances, the~class-wise accuracies varied largely between 41.5\% ({peas}) and 96.8\% ({maize}).
For better visibility, we emphasized the best and worst metrics by boldface.
The conditional kappa scores are similarly variable and range between $0.414$ ({peas}) and $0.957$ ({rapeseed}).

Frequent classes (e.g., {maize}, {meadow}) have been in general more confidently classified than less frequent classes (e.g., {peas}, {summer oat}, {winter spelt}, {winter triticale}).
Nonetheless this relation has exceptions.
The least frequent class, {peas}, performed relatively well on data of 2016, and~other less frequent classes, such as {asparagus} or {hop}, showed good performances despite their underrepresentation in the dataset.

\newtext{
	To investigate the causes of the varying accuracies, we calculated confusion matrices for both seasons as shown in \cref{fig:confmat}.
	These error matrices are two-dimensional histograms of classification samples aggregated by the class prediction and ground truth reference.
	To account for the non-uniform class distribution, the absolute number of samples for each row-column pair is normalized.
	We~decided to normalize the confusion matrices by row to obtain recall (producer's) accuracies, due~to their direct relation to available ground truth labels.
	The diagonal elements of the matrices represent correctly-classified samples with values equivalent to \cref{tab:accuracies}.
	Structures outside the diagonal indicate systematic confusions between classes and may give insight into the reasoning behind varying classification~accuracies.}

Some crops likely share common spectral or phenological characteristics.
Hence, we expected some symmetric confusion between classes, which would be expressed as diagonal symmetric confusions consistent in both years.
Examples of this were {triticale} and {rye} or {oat} and {summer barley}.
However, these relations were not frequent in the dataset, which indicates that the network had sufficient capacity to separate the classes by provided features.
In some cases, one class may share characteristics with another class.
This class may be further distinguished by additional unique features, which would be expressed by asymmetric confusions between these two classes in both seasons.
Relations of this type were more dominantly visible in the matrices and included confusions between {barley} and {triticale}, {triticale} and {spelt} or {wheat} confused with {triticale} and {spelt}.
These types of confusion were consistent over both seasons and may be explained by a spectral or phenological similarity between individual crop-types.

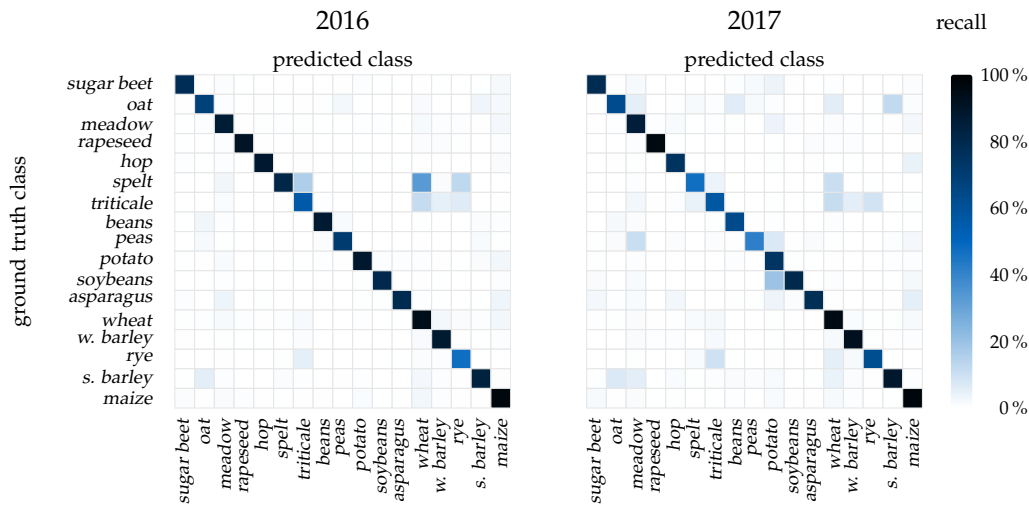
\begin{figure}
	\centering
	
	\tikzsetnextfilename{confmatboth}
	
	\def\dataindex{4}
	\def\cbartitle{recall}
	


\begin{tikzpicture}

  \pgfplotsset{every axis label/.append style={font=\footnotesize},tick pos=right, ylabel near ticks}
  
  \pgfplotsset{
    axis line style={%
      opacity=0 
    }   
  }
  


  \begin{groupplot}[
  	group style={
  		group size=2 by 1,
  		xlabels at=edge bottom,
  		ylabels at=edge left,
  		xticklabels at=edge bottom,
  		vertical sep=35pt,
  		group name=seq_len_plot
  	},
  	title style={yshift=.75em,},
    width=6cm,
    height=6cm,
    enlargelimits=false,
    xtick=data,
    xtick distance=1,
    ytick distance=1,
    colormap={example}{%
		color=(tumwhite)
		color=(tumblue)
		color=(tumblack)
	},
    ytick=data,
    ytick align=outside,
    tick style={draw=none},
    ytick pos = left,  
    yticklabel style = {yshift=-0mm, rotate=0,anchor=east,font=\scriptsize},
    xticklabel style = {yshift=-0mm, rotate=90,font=\scriptsize},
    yticklabel pos=left,
    xlabel={predicted class},
    x label style={at={(axis description cs:0.5,1.1)},anchor=south},
    y label style={at={(axis description cs:-0.4,.5)},anchor=south},
    ylabel={ground truth class},
    xticklabels={},
  ]
  \nextgroupplot[title=2016,
      yticklabels={
      	\classname{sugar beet},
      	\classname{oat},
      	\classname{meadow},
      	\classname{rapeseed},
      	\classname{hop},
      	\classname{spelt},
      	\classname{triticale},
      	\classname{beans},
      	\classname{peas},
      	\classname{potato},
      	\classname{soybeans},
      	\classname{asparagus},
      	\classname{wheat},
      	\classname{w. barley},
      	\classname{rye},
      	\classname{s. barley},
      	\classname{maize}
      },
      xticklabels={
      	\classname{sugar beet},
      	\classname{oat},
      	\classname{meadow},
      	\classname{rapeseed},
      	\classname{hop},
      	\classname{spelt},
      	\classname{triticale},
      	\classname{beans},
      	\classname{peas},
      	\classname{potato},
      	\classname{soybeans},
      	\classname{asparagus},
      	\classname{wheat},
      	\classname{w. barley},
      	\classname{rye},
      	\classname{s. barley},
      	\classname{maize}
      },
  ]
  
    \addplot[
      matrix plot,
        shader=faceted,
        faceted color=tumgraylight,
      mesh/cols=17,
      empty line=scanline,
      point meta=explicit,
      point meta min=0,
      point meta max=1,
    ] table[meta index=\dataindex] {images/confmat/formatted_grucm2016.csv};
    
   \nextgroupplot[
       title=2017,
       colorbar right,
       colorbar style={
	       	title={\cbartitle}, 
	       	font=\footnotesize,
	       	anchor=north west,
	       	width=8pt,
	       	ytick={
	       		0,
	       		.2,
	       		.4,
	       		.6,
	       		.8,
	       		1
       		},
	       	yticklabels={
	       		\SI{0}{\percent},
	       		\SI{20}{\percent},
	       		\SI{40}{\percent},
	       		\SI{60}{\percent},
	       		\SI{80}{\percent},
	       		\SI{100}{\percent}
       		},
	       	ticklabel pos=right,
	       	ticklabel style={
	       		anchor=east,
	       		xshift=2em,	
	       		align=left
       		},
	       	label style={yshift=1em},
	       	rounded corners=1pt
	       },
       yticklabels={},
     xticklabels={
      	\classname{sugar beet},
		\classname{oat},
		\classname{meadow},
		\classname{rapeseed},
		\classname{hop},
		\classname{spelt},
		\classname{triticale},
		\classname{beans},
		\classname{peas},
		\classname{potato},
		\classname{soybeans},
		\classname{asparagus},
		\classname{wheat},
		\classname{w. barley},
		\classname{rye},
		\classname{s. barley},
		\classname{maize}
		     },
       ]
    
    \addplot[
    matrix plot,
    ,   
    shader=faceted,
    faceted color=tumgraylight,
    mesh/cols=17,
    empty line=scanline,
    point meta=explicit,
    point meta min=0,
    point meta max=1,
    ] table[meta index=\dataindex] {images/confmat/formatted_grucm2017.csv};
    
  \end{groupplot}
  
\end{tikzpicture}

	\caption{%
		Confusion matrix of the trained convolutional \gls{gru} network on data of the seasons 2016 and 2017.
		While the confusion of some classes was consistent over both seasons (\eg \classname{winter triticale} to \classname{winter wheat}), other classes are  classified at different accuracies at consecutive years (\eg \classname{winter barley} to \classname{winter spelt}).
	}
	\label{fig:confmat}
\end{figure}


}

More dominantly, many confusions were not consistent over the two growing seasons.
For~instance, confusions occurring only in the 2017 season were {soybeans} with {potato} or {peas} with {meadow} and {potato}.
Since the cultivated crops are identical in these years and the class distributions were consistent, seasonally-variable factors were likely responsible for these relations.
As reported in \cref{tab:accuracies}, {peas} have been classified well in 2016, but poorly in 2017, due to the aforementioned confusions with {meadow} and {potato}.
These results indicate that external and not crop-type-related factors had a negative influence on classification accuracies, which appeared unique to one season.
One of these might be the variable onset of phenological events, which are indirectly observed by the change of reflectances by the sensors.
These events are influenced by local weather and sun exposure, which may vary over large regional scales or multiple years.



\subsection{Qualitative Classification Evaluation}
\label{sec:qualitative}


\begin{figure}[t]
  \input{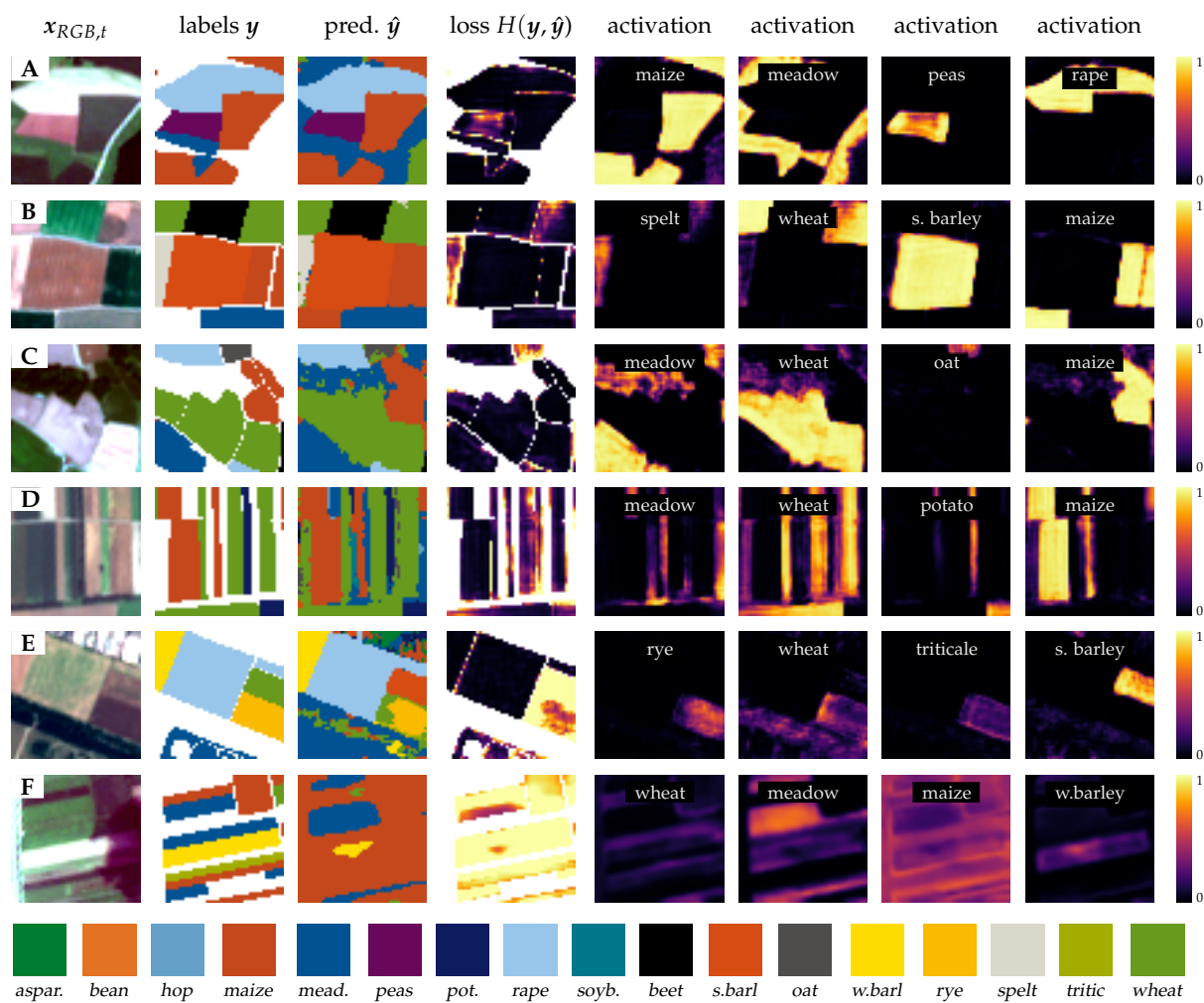}
  \caption{
  	Qualitative results of the convolutional \gls{gru} sequential encoder. 
  	Examples \textsc{A-D} show good classification results.
  	At example \textsc{E} the network misclassified one \cn{maize} parcel with high confidence, which is indicated by incorrect, but well defined activations.
  	At a second field the class activations reveal a confusion between \cn{wheat}, \cn{meadow} and \cn{maize}.
    At example \textsc{F} most pixels are misclassified. 
    However, the class activations show uncertainty in the classification decision.
  }
  \label{fig:examples}
\end{figure}

For the qualitative evaluation, we used the same network structure as in the previous section.
We~inferred the network with \SI{48}{\pixel} tiles from the evaluation dataset of 2017 for better visibility.
In \cref{fig:examples}, a series of good ({A--D}) and bad ({E},{F}) classification examples are shown.
The first column represents the input sequence $\VInput$ as band-normalized RGB images from one selected cloud-free observation $\VInput_{RGB,t}$.
Further columns show the available ground truth labels $\V{y}$, predictions $\hat{\V{y}}$ and the cross-entropy loss $H(\V{y},\hat{\V{y}})$.
Additionally, four selected softmax-normalized class activations are displayed in the last columns.
These activations can be interpreted as classification confidences for each class.
The~prediction map contains the index of the most activated class at each pixel, which may be interpreted as the class of highest confidence.
The cross-entropy loss is the measure the agreement between the one-hot representation of the ground truth labels and the activations per class.
It is used as the objective function, as network training indicates disagreement between ground truth and prediction even if the final class prediction is correct.
This relation can be observed in fields of several examples, such as {peas} in Example {A}, {spelt} in Example {B} and {oat} in Example {C}.
However, most classifications for these examples were accurate, which is expressed by well-defined activation maps.

\newtext{
	Often, classifiers use low-pass filters in the spatial dimensions to compensate for high-frequent noise.
	These filters typically limit the ability to classify small objects.
	To evaluate to what degree the network has learned to apply low-pass filtering, we show a tile with a series of narrow fields in Example {D}.
}
Two thin {wheat} and {maize} fields have been classified correctly.
However, some errors occurred on the southern end of an adjacent {potato} field, as indicated by the loss map.
It appears that the network was able to resolve high-frequency spatial changes and did not apply smoothing of the class activations, as in Example {F}.
%



Two misclassified fields are shown in Example {E}.
The upper {wheat} field has been confidently misclassified to {summer barley}.
Underneath, the classification of a second {rye} field was uncertain between {rye}, {wheat} and {triticale}.
While {triticale}, as the least activated class, was not present in the prediction map, the mixture of {rye} and {wheat} is visible in the class predictions.

\textls[-5]{Example {F} shows a mostly misclassified tile.
	Only a few patches of {meadow} and {winter barley} were correctly predicted.
	The activations of these classes were, compared to previous examples, generally more blurred and of lower amplitude.
	Similar to Example {D}, the most activated classes are also the most frequent in the dataset.
	In fact, the entire region around the displayed tile seemed to be classified poorly.
	This region was located on the northwest border of the \gls{aoi}.
	Further~examination showed that for this region, fewer satellite images were available.
	The lack of temporal information likely explains the poor classification accuracies.
	However, this example illustrates that the class activations give an indication of the classification confidence independent of the ground truth information.}

\section{Discussion}
\label{sec:discussion}

\glsresetall

In this section, we compare our approach with other multi-temporal classifications.
\newtext{
	Unfortunately, to the best of our knowledge, no multi-temporal benchmark dataset is available to compare remote sensing approaches on equal footing.
}
Nevertheless, we provide some perspective of the study domain by gathering multi-temporal crop classification approaches in \cref{tab:approaches} and categorizing these by their applied methodology and achieved overall accuracy.
However, the heterogeneity of data sources, the~varying extents of their evaluated areas and the number of classes used in these studies impedes a numerical comparison of the achieved accuracies.
Despite this, we hope that this table will provide an overview of the state-of-the-art in multi-temporal crop identification.

\textls[-15]{\Gls{eo} data are acquired in periodic intervals at high spatial resolutions.
	From~an information theoretical perspective, utilizing additional data should lead to better classification performance.
	However, the large quantity of data requires methods that are able to process this information and are robust with regard to observation noise.
	Optimally, these approaches are scalable with minimal supervision so that data of multiple years can be included over large regions.
	Existing~approaches in multi-temporal \gls{eo} tasks often use multiple separate processing steps, such~as preprocessing, feature extraction and classification, as summarized by~\citet{Uensalan11:RLU}.
	Generally, these steps require manual supervision or the selection of additional parameters based on region-specific expert knowledge, a process that impedes applicability at large scales.}
The cost of data acquisition is an additional barrier, as multiple and potentially expensive satellite images are required.
Commercial satellites, such~as \acrfull{rapideye}, \acrfull{spot} or~\acrfull{quickbird}, provide images at excellent spatial resolution.
However, predominantly inexpensive sensors, such as \gls{ls}, \gls{s2}, \gls{modis} or \gls{aster}, can be applied at large scales, since the decreasing information gain of additional observations must justify image acquisition costs.
Many approaches use spectral indices, such as \gls{ndvi}, \gls{ndwi} or \gls{evi}, to extract statistical features from vegetation-related signals and are invariant to atmospheric perturbations.
Commonly, \glspl{dt} or \glspl{rf} are used for classification.
The exclusive use of spectral indices simplifies the task of feature extraction.
However, these indices utilize only a small number of available spectral bands (predominantly {blue}, {red}~and {near-infrared}).
Thus, methods that utilize all reflectance measurements, either at \gls{toa}, or atmospherically-corrected to \gls{boa}, are favorable, since all potential spectral information can be extracted.

\begin{table}[]
  \centering
  \caption{%
    Overview over recent approaches for crop classification.
  }
  \label{tab:approaches}
  \arrayrulecolor{tumgray}
\newcommand{\atcor}{atm. cor. }

\begin{tabular}{@{}
                X
                v{11mm}
                v{28mm}
                v{20mm}
                v{15mm}
                v{10mm}
                p{10mm}
                @{}}
\toprule
\multicolumn{1}{@{}X}{Approach} & \multicolumn{6}{X@{}}{Details} \\
\cmidrule(l){2-7}

& 
\multicolumn{1}{X}{Sensor} &
\multicolumn{1}{X}{Preprocessing} &
\multicolumn{1}{X}{Features} & 
\multicolumn{1}{X}{Classifier} & 
\multicolumn{1}{X}{Accuracy} &
\multicolumn{1}{X@{}}{\# Classes} \\ 

\cmidrule(r){1-1}
\cmidrule(lr){2-2}
\cmidrule(lr){3-3}
\cmidrule(lr){4-4}
\cmidrule(lr){5-5}
\cmidrule(lr){6-6}
\cmidrule(l){7-7}
\addlinespace
{this work} & 
\acrshort{s2} & 
none & 
\acrshort{toa} reflect. &
\acrshort{convrnn} & 
90 &
17 \\
\addlinespace
\addlinespace
\citet[\citeyear{Russwurm17:TVM}]{Russwurm17:TVM} & 
\acrshort{s2} & 
\atcor (\brand{sen2cor}) & 
\acrshort{boa} reflect. &
\acrshort{rnn} & 
74 & 
18 \\ 
\addlinespace
{\citet[\citeyear{Siachalou2015}]{Siachalou2015}} & 
\acrshort{ls}, \acrshort{rapideye} & 
geometric correction,\newline
image registration & 
\acrshort{toa} reflect. &
\acrshort{hmm} & 
90 & 
6 \\ 
\addlinespace
\citet[\citeyear{Hao15}]{Hao15} &
\acrshort{modis} & 
image reprojection,\newline \atcor \cite{Richter1996} & 
statistical phen. features &
\acrshort{rf} & 
89 &
6 \\ 
\addlinespace
\citet[\citeyear{conrad2014}]{conrad2014} &
\acrshort{spot},\newline \acrshort{rapideye}, \acrshort{quickbird} & 
segmentation, \newline \atcor  \cite{Richter1996} & 
vegetation indices &
\acrshort{obia}+\acrshort{rf} & 
86 &
9 \\ 
\addlinespace
\citet[\citeyear{Foerster2012}]{Foerster2012} &
\acrshort{ls} & 
phen. normalization,\newline \atcor \cite{Richter1996} & 
\acrshort{ndvi} statistics &
\acrshort{dt}& 
73 &
11 \\ 
\addlinespace
\citet[\citeyear{Barragan2011}]{Barragan2011} &
\acrshort{aster} & 
segmentation,\newline \atcor \cite{Matthew2000} & 
vegetation indices &
\acrshort{obia}+\acrshort{dt} & 
79 &
13 \\ 
\addlinespace
\citet[\citeyear{conrad2010}]{conrad2010} &
\acrshort{spot}\newline\acrshort{aster} & 
segmentation,\newline\atcor \cite{Richter1996} & 
vegetation indices &
\acrshort{obia}+\acrshort{dt}& 
80 &
6 \\ 
\addlinespace
\bottomrule
\end{tabular}       

\end{table}

In general, a direct numerical comparison of classification accuracies is difficult, since these are dependent on the number of evaluated samples, the extent of evaluated area and the number of classified categories.
\textls[-12]{Nonetheless, we compare our method with the approaches of \mbox{\citet{Siachalou2015}} and~\citet{Hao15} in detail since their achieved classification accuracies are on a similar level as ours.
	\mbox{\Citet{Hao15}} used an \gls{rf} classifier on phenological features, which were extracted from \gls{ndvi} and \gls{ndwi} time series of \gls{modis} data.
	Their results demonstrate that good classification accuracies with hand-crafted feature extraction and classification methods can be achieved if data of sufficient temporal resolution are available.
	However, the large spatial resolution (\SI{500}{\meter}) of the \gls{modis} sensor limits the applicability of this approach to areas of large homogeneous regions.}
On a smaller scale, \mbox{\citet{Siachalou2015}} report good levels of accuracy on small fields.
For this, they used a \glspl{hmm} with a temporal series of four \gls{ls} images combined with one single \gls{rapideye} image for field border delineation. 
Methodologically, \glspl{hmm} and \glspl{crf}~\citep{Hoberg2015:CRF} are closer to our approach since the phenological model is approximated with an internal chain of hidden states.
However, these~methods might not be applicable for long temporal series, since Markov-based approaches assume that only one previous state contains classification-relevant information.

Overall, this comparison shows that our proposed network can achieve state-of-the-art classification accuracy with a comparatively large number of classes.
Furthermore, the \gls{s2} data of non-atmospherically-corrected \gls{toa} values can be acquired easily and does not require further preprocessing.
Compared to previous work, we were able to process larger tiles by using convolutional recurrent cells with only a single recurrent encoding layer.
Moreover, we neither required atmospheric correction, nor additional cloud classes, since one classification decision is derived from the entire sequence of observations.

\section{Conclusion}

In this work, we proposed an automated end-to-end approach for multi-temporal classification, which achieved state-of-the-art accuracies in crop classification tasks with a large number of crop classes.
Furthermore, the reported accuracies were achieved without radiometric and geometric preprocessing.
The trained and inferred data were atmospherically uncorrected and contained clouds.
In traditional approaches, multi-temporal cloud detection algorithms utilize the sudden positive change in reflectivity of cloudy pixels and achieve better results than other traditional mono-temporal remote sensing classifiers~\cite{Hagolle10}. 
Results of this work indicate that cloud masking can be learned jointly together with classification.
By visualizing internal gate activations in our network in \cref{sec:activations}, we~found evidence that some recurrent cells were sensitive to cloud coverage.
These cells may be used by the network to internally mask cloudy pixels similar to an external cloud filtering algorithm.

In \cref{sec:quantitative,sec:qualitative}, we further evaluated the classification results quantitatively and qualitatively.
Based on several findings, we derived that the network has approximated a discriminative crop-specific phenological model based on a raw series of \gls{toa} \gls{s2} observations.
Further inspection revealed that some crops were inconsistently classified in both growing seasons.
This may be caused by seasonally-variable environmental conditions, which may have been implicitly integrated into the encoded phenological model.
\newtext{
	We employed our network for the task crop classification since vegetative classes are well characterized by their inherently temporal phenology.
	However, the~network architecture is methodologically not limited to vegetation modeling and may be employed for further tasks, which may benefit from the extraction of temporal features.
}
We hope that our results encourage the research community to utilize the temporal domain for their applications.
In this regard, we publish the \brand{TensorFlow} 
source code of our network along with the evaluations and experiments from this work.

\supplementary{The source code of the network implementation and further material is made publicly available at \url{https://github.com/TUM-LMF/MTLCC}.
}

\acknowledgments{We would like to thank the \acrfull{stmelf}
	for providing ground truth data in excellent semantic and geometric quality.
	Furthermore, we thank the \gls{lrz}
	for providing access to computational resources, such as the DGX
	-1 and P100
	servers and \brand{Nvidia} for providing one \brand{Titan X GPU}.}

\authorcontributions{M.R. and M.K. conceived and designed the experiments. M.R implemented the network and performed the experiments. Both authors analyzed the data and M.R. wrote the paper. Both authors read and approved the final manuscript.}

\conflictsofinterest{The authors declare no conflict of interest.} 

\section*{ArXiv version history}
\begin{description}
	\item[v1] submitted version to IJGI
	\item[v2] added notice of submission date to history
	\item[v3] revised manuscript version on review comments. Added section \cref{sec:prevwork} \textsl{Prior Work} and improved English language and style. 
	\item[v4] updated ArXiv version with published MDPI paper and added doi. Minor spelling and wording corrections and text layout. Minor modifications on cited papers. Updated Github link.
\end{description}

\externalbibliography{yes}
{
\bibliography{bib/IJGI2017}
}


\end{document}